\documentclass[10pt,journal,compsoc]{IEEEtran}
\ifCLASSOPTIONcompsoc
  \usepackage[nocompress]{cite}
\else
  \usepackage{cite}
\fi

\usepackage{amsmath,amssymb,amsfonts}
\usepackage{bm}
\usepackage{dsfont}
\usepackage{graphicx}
\usepackage{adjustbox}
\usepackage[bookmarks=false]{hyperref}
\usepackage{comment}
\usepackage{algorithm}
\usepackage{algpseudocode}
\usepackage{units}
\usepackage{multirow}
\usepackage{array}
\usepackage{booktabs}
\usepackage{svg}
\usepackage{color}
\usepackage{cite}
\usepackage{threeparttable, tablefootnote}
\usepackage{adjustbox}
\usepackage{enumitem}
\usepackage[doipre={DOI:\ }]{uri}
\setlist[itemize]{align=parleft,left=0pt..1em}

\definecolor{darkspringgreen}{rgb}{0.09, 0.45, 0.27}

\usepackage[absolute,showboxes]{textpos}

\TPMargin{5pt}

\newcommand{\copyrightstatement}{
    \begin{textblock}{0.88}(0.06,0.95)
         \noindent
         \scriptsize \textcopyright 2024 IEEE. Personal use of this material is permitted.  Permission from IEEE must be obtained for all other uses, in any current or future media, including reprinting/republishing this material for advertising or promotional purposes, creating new collective works, for resale or redistribution to servers or lists, or reuse of any copyrighted component of this work in other works. Accepted as a journal paper for IEEE Transactions on Computers.~\doi{10.1109/TC.2024.3449084}
    \end{textblock}
}

\begin{document}
\bstctlcite{IEEEexample:BSTcontrol}
\title{Joint Pruning and Channel-wise Mixed-Precision Quantization for Efficient Deep Neural Networks}
\author{Beatrice~Alessandra~Motetti,
        Matteo~Risso,~\IEEEmembership{Student Member,~IEEE,}
        Alessio~Burrello,~\IEEEmembership{Member,~IEEE,}
        Enrico~Macii,~\IEEEmembership{Fellow,~IEEE,}
        Massimo~Poncino,~\IEEEmembership{Fellow,~IEEE,}
        and~Daniele~Jahier~Pagliari,~\IEEEmembership{Member,~IEEE}%
\IEEEcompsocitemizethanks{\IEEEcompsocthanksitem B.A. Motetti, M. Risso, A. Burrello, E. Macii, M. Poncino and D. Jahier Pagliari are with Politecnico di Torino, 10129, Turin, Italy. E-mail: firstname.firstsurname@polito.it
\IEEEcompsocthanksitem This publication is part of the project PNRR-NGEU which has received
funding from the MUR – DM 118/2023. This work has received funding from the Key Digital Technologies Joint Undertaking (KDT-JU) under grant agreements No 101095947 and No 101112274. The JU receives support from the European Union’s Horizon Europe research and innovation programme.
\IEEEcompsocthanksitem We acknowledge the CINECA award under the ISCRA initiative, for the availability of high performance computing resources and support.
}

}


\IEEEtitleabstractindextext{%
\begin{abstract}
The resource requirements of deep neural networks (DNNs) pose significant challenges to their deployment on edge devices. 
Common approaches to address this issue are pruning and mixed-precision quantization, which lead to latency and memory occupation improvements.
These optimization techniques are usually applied independently.
We propose a novel methodology to apply them jointly via a lightweight gradient-based search, and in a hardware-aware manner, greatly reducing the time required to generate Pareto-optimal DNNs in terms of accuracy versus cost (i.e., latency or memory).
We test our approach on three edge-relevant benchmarks, namely CIFAR-10, Google Speech Commands, and Tiny ImageNet. When targeting the optimization of the memory footprint, we are able to achieve a size reduction of 47.50\% and 69.54\% at iso-accuracy with the baseline networks with all weights quantized
at 8 and 2-bit, respectively.
Our method surpasses a previous state-of-the-art approach with up to 56.17\% size reduction at iso-accuracy.
With respect to the sequential application of state-of-the-art pruning and mixed-precision optimizations, we obtain comparable or superior results, but with a significantly lowered training time.
In addition, we show how well-tailored cost models can improve the cost versus accuracy trade-offs when targeting specific hardware for deployment.
\end{abstract}

}

\maketitle
\copyrightstatement

\IEEEdisplaynontitleabstractindextext
\IEEEpeerreviewmaketitle

\ifCLASSOPTIONcompsoc
\IEEEraisesectionheading{\section{Introduction}\label{sec:introduction}}
\else
\section{Introduction}
\label{sec:introduction}
\fi
\IEEEPARstart{D}{eep} neural networks (DNNs) have showcased impressive performance in a wide range of domains; nevertheless, their computational complexity often clashes with 
the resource limitations of hardware (HW) devices, especially at the edge~\cite{liberis2021micronas}. 
Thus, several studies have explored techniques aimed at reducing the computational complexity and memory requirements of DNNs, while preserving their task performance. 
Pruning\cite{han2015efficient} and quantization\cite{jacob2018quantization} are two of the most popular ones.
The former eliminates unimportant computations from a network, reducing the number of parameters and operations, while quantization, in particular with Mixed-Precision Search (MPS), optimizes the data representation of the model's parameters and activations.

\begin{figure}[]
  \centering
  \includegraphics[width=\columnwidth]{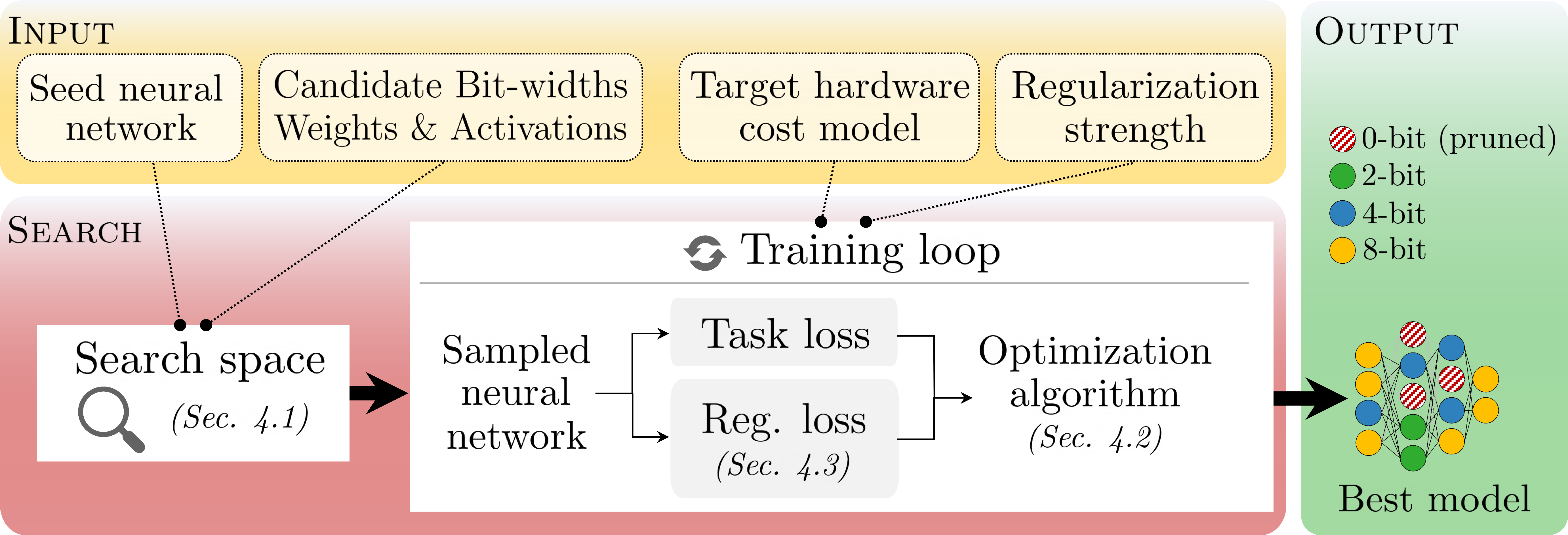}
      \vspace{-0.7cm}
      \caption{Overview of the key components of the proposed method, with the required inputs (yellow), the core optimization scheme (red), and the final result, i.e. a pruned mixed-precision DNN (green).}
  \label{fig:method_overview}
  \vspace{-0.45cm}
\end{figure}

Finding the optimal pruning rate for different parts of a network, as well as the optimal quantization precision, involves solving a complex optimization problem. Early approaches used black-box optimization methods, such as Reinforcement Learning~(RL) and Evolutionary Algorithms~(EA)~\cite{white2023nas1000papers, benmeziane2021hwawarenas}. While effective, these methods are based on time-consuming iterative procedures. 
Motivated by this, gradient-based methods (also known as \textit{one-shot}) have emerged as lightweight yet competitive alternatives, for both pruning and MPS~\cite{risso2023pit, cai2020edmips}. These techniques simultaneously train the quantization or pruning configurations together with the weights of the architecture, leading to convergence in a time comparable to a single training.
Pruning and MPS are usually applied independently (e.g., one after the other), and most of the approaches that consider them jointly use time-consuming black-box methods.

In this paper, we extend the gradient-based precision assignment technique proposed by Risso \textit{et al.}~\cite{risso2022igsc}, which assigns the bit-width to each individual channel of the weights of the DNN, enabling it to \textit{jointly perform pruning and MPS} in a differentiable manner.
To our knowledge, ours is the first gradient descent-based optimization method to consider both \textit{channel-wise} MPS and pruning simultaneously.
This eliminates the need of a time-consuming concatenation of optimizations, which might also unnecessarily restrict the search space, limiting the choices available to the second method applied (e.g., MPS) to the results of the first one (e.g., pruning). Additionally, ours is also the first one-shot method to include hardware-aware inference cost models in the optimization, showing their importance to obtain models tailored to the deployment target.
An overview of our proposed method is shown in Fig.~\ref{fig:method_overview}. 
In detail, the main novel contributions of this work, that specifically targets tiny convolutional neural networks to be deployed at the edge, are the following:
\begin{itemize}
    \item We propose a gradient-based approach to jointly explore mixed-precision quantization and pruning. In particular, we explore a channel-wise weights' precision assignment, extending our previous work~\cite{risso2022igsc} with the option to quantize at "0 bits", which corresponds to pruning the weights' channels of a layer in a structured way.
    \item 
    To quantify the complexity of the networks explored by our tool, we consider different cost models, both hardware-agnostic (targeting model size reduction) and hardware-specific (targeting latency reduction). We build latency models for two real-world mixed-precision enabled hardware targets, i.e. the Mixed Precision Inference Core~(MPIC)~\cite{ottavi2020mpic} and the Neural Engine 16~(NE16) DNN accelerator~\cite{macan2023ne16}, and analyze the impact of defining the correct cost model when targeting a specific device.
    \item We evaluate our proposed approach on three edge-relevant benchmarks, namely CIFAR-10, Google Speech Commands v2~\cite{google-speech-commands}  and Tiny ImageNet. 
    We obtain size reductions up to 47.50\% at iso-accuracy with baseline networks with all weights quantized at 8 bits, and up to 69.54\% at iso-accuracy with 2-bit models.
    Furthermore, our models are up to 56.17\% smaller in size than those obtained with a state-of-the-art MPS method~\cite{risso2022igsc}.
    We achieve comparable performances with respect to the concatenation of channel-wise pruning, using PIT~\cite{risso2023pit}, and channel-wise precision assignment proposed by Risso \textit{et al.}~\cite{risso2022igsc}, but our joint method is significantly faster.
\end{itemize}

Our code is open-source at: \texttt{\url{https://github.com/eml-eda/mixprec-pruning}}.
The paper is structured as follows. Sec.~\ref{sec:background} covers the required background concepts, and Sec.~\ref{sec:related} summarizes the most relevant works on mixed-precision quantization and pruning. In Sec.~\ref{sec:methods} we illustrate our proposed approach and in Sec.~\ref{sec:results} the experimental results. Lastly, Sec.~\ref{sec:conclusion} concludes the paper.

\section{Background}
\label{sec:background}
\subsection{Quantization and Mixed-Precision Search}~\label{subsec:quantization}
Quantization is one of the key optimizations to reduce the complexity of DNNs.
In particular, integer quantization is extensively used to replace the floating-point representation of both weights and activations with low bit-width integers.
This leads to model compression, and to the usage of integer arithmetic operators, which are faster and more energy-efficient~\cite{jacob2018quantization}.
Moreover, integer quantization also enables the execution of DNNs even on edge devices without a floating point unit.
Quantization can be applied on networks already trained in float (i.e., post-training quantization), or its effect can be simulated during training (i.e., quantization-aware training, which is often beneficial for accuracy~\cite{jacob2018quantization}.

This work considers the well-known \textit{affine quantization} scheme, that maps float tensors $\mathbf{T}$ to $n$-bit integers as:
\begin{equation}
    \mathbf{T}_n = \underset{0:2^{n}-1}{\mathrm{clamp}}\left( \mathrm{round}\left( \frac{\mathbf{T} - \alpha_\mathbf{T}}{\varepsilon_\mathbf{T}} \right)\right) \label{eq:quantz}
    \vspace{-0.1cm}
\end{equation}
where $\mathrm{clamp}$ restricts the values to the interval $[0:2^{n}-1]$, $[\alpha_\mathbf{T},\beta_\mathbf{T})$ is the range of values that can be mapped without saturation, and $\varepsilon_\mathbf{T} = (\beta_\mathbf{T}-\alpha_\mathbf{T}) / (2^{n}-1)$ is the quantization step.
In general, $\alpha_\mathbf{T}$ and $\beta_\mathbf{T}$ can be set independently for each \textit{layer}, \textit{channel}, or \textit{block} of weights or activations~\cite{dai2021vsquant, frantar2022gptq}. While layer-wise and channel-wise assignments are substantially equivalent, block-wise quantization incurs much higher memory overheads, and requires substantial changes to the layers' execution flow (needing a separate rescaling for each block of weights). Thus, it is not compatible with most DNN accelerators, or inference libraries for general-purpose HW. Accordingly, in our work, we use per-channel quantization parameters everywhere.
Affine quantization has been explored in works such as PACT~\cite{choi2018pact} and LQ-Nets~\cite{lq_nets}, which also \textit{learn} the value range and the optimal step during training.
Note that our proposed algorithm is orthogonal to the specific quantization schemes. Still, we test it on affine quantization due to its hardware friendliness.

Conventional quantization uses \textit{fixed-precision}, wherein although different parts of a DNN can use different quantization parameters, a uniform bit-width, typically 8 bits, is applied throughout the entire model.
Recently, however, \textit{mixed-precision} strategies have gained attention~\cite{cai2020edmips}. They involve using different bit-widths for different sections of a DNN, resulting in additional optimization opportunities in terms of time, memory, and energy consumption, especially when the underlying hardware natively supports operations at sub-byte precision~\cite{ottavi2020mpic, macan2023ne16}. Nonetheless, determining the optimal allocation of bit-widths to different segments of the network presents a substantial challenge, as it entails exploring a large solution space that grows exponentially with the DNN's depth.
Recently, approaches inspired by Neural Architecture Search (NAS)~\cite{nas_rl, liu2018darts} have been applied to solve this so-called Mixed-Precision Search problem. 
Such approaches can be classified into two families, namely \textit{iterative} and \textit{one-shot}.

Iterative approaches leverage black-box optimization engines like RL~\cite{haq,releq}. They entail the iterative sampling of one or more designs from the search-space, followed by an evaluation step that consists in a complete training, for evaluating accuracy, and a deployment on the target (or the use of a proxy model) for measuring non-functional metrics such as latency and energy. Then, the RL agent is updated in light of the evaluation's outcomes, and the cycle is repeated. 
These schemes are flexible and versatile, but scale poorly with the dimension of the search-space requiring thousands of GPU hours for a single search~\cite{tan2019mnasnet}.

Conversely, one-shot approaches require a single joint precision search and training loop, thus solving the main issue of iterative MPS.
This is achieved at the price of having an analytical differentiable form of the optimization objective which allows to jointly train weights and explore bit-widths assignments with gradient-descent. For this reason, one-shot methods are also referred to as \textit{gradient-based} or \textit{differentiable}.
Both iterative and one-shot state-of-the-art MPS schemes will be revised in Sec.~\ref{subsec:mpq}.

\subsection{Pruning}~\label{subsec:pruning}
Orthogonal to quantization, another key optimization for DNN compression and inference speed-up is pruning~\cite{pruning_survey}. While quantization works at the operand representation level, pruning involves reducing the size of DNNs by selectively removing a subset of their parameters.
Pruning techniques can be classified based on the granularity of the subsets to be removed and on the selection scheme employed to decide which subsets to eliminate.

According to the pruning granularity, two main approaches can be identified, i.e., \textit{unstructured} and \textit{structured}.
Unstructured pruning considers the removal of single weights.
This approach leads to the largest theoretical reduction of parameters and operations without accuracy loss, reaching values as high as 90\%~\cite{deep_compression}. However, it leads to sparsely connected networks that cannot be easily accelerated in hardware, especially on general-purpose platforms, without adding a significant overhead~\cite{scalpel,pruning_survey}. In fact, sparse computations make it complex to achieve high utilization on parallel hardware and damage the regularity and locality of memory access patterns, resulting in worse cache behaviour~\cite{scalpel}.

Structured pruning refers to those approaches that remove weights with some regularity, e.g., in blocks, or that eliminate entire neurons, channels or filters. While block-wise and block-balanced pruning still require hardware support to be effectively accelerated~\cite{pruning_survey}, channel-/filter-wise pruning leads to compressed architectures where the associative and distributive properties of linear algebra can be used to transform them into smaller dense structures~\cite{pruning_survey}, thus solving the main drawback of the unstructured approaches. The deployment of the obtained networks is straightforward and does not incur any overhead, nor needs special hardware/software support.
The main disadvantage is represented by the worse achievable trade-off between task performance degradation and compression~\cite{deep_compression}.

For both unstructured and structured pruning techniques, different selection schemes have been proposed in literature.
Removing weights with the lowest absolute magnitude is a simple yet effective strategy~\cite{deep_compression, data_free_struct}. It works for both individual weights and weight groups~\cite{data_free_struct}, and it can be applied without the need for additional data. However, highly compressed networks may experience reduced performance, requiring additional fine-tuning and potentially undermining the data-free benefit~\cite{pruning_survey}.
To solve this issue, sensitivity-based methods employ training data to determine which elements to remove based on their impact on the output when exposed to different input examples~\cite{thinet}.
However, sensitivity-based methods are still applied to a network already trained at convergence (possibly fine-tuned afterwards). Thus, they do not exploit the potential optimization opportunities offered by jointly training and pruning the weights~\cite{pruning_survey}. Gradient-based methods, instead, do exactly that, allowing the training optimizer to progressively tune the network's weights in order to recover the accuracy drops caused by pruning~\cite{morphnet,risso2023pit}. Typically, this is achieved by enhancing standard DNNs with additional trainable binary gates that control which portions of the DNN to prune. Both the standard DNN's weights and the gates are then jointly trained with gradient-descent to minimize a loss composed of the task loss and cost-related objectives (e.g., number of parameters). 

In this work, we consider a structured pruning scheme with learnable binary gates from this last category, which allows to easily achieve effective speed-up and model compression when deployed on the actual HW, while preserving the task performance.

\section{Related Works}
\label{sec:related}
\subsection{Mixed-Precision Search} \label{subsec:mpq}

\begin{table*}[]
\centering
\begin{adjustbox}{max width=\textwidth}
\begin{threeparttable}
\caption{Summary of most similar state-of-the-art Mixed-Precision Search and pruning methods}
\vspace{-0.2cm}
\label{tab:related}
\begin{tabular}{llllll}
\hline
\textbf{Method} & \textbf{Search Engine} & \textbf{HW-awareness}\tnote{a} & \textbf{Supported Bit-widths} & \textbf{Per-Channel Quantization} & \textbf{Pruning} \\ \hline
HAQ~\cite{haq} & RL & High & Any & No & None \\
ReLeQ~\cite{releq} & RL & High & Any & No & None \\
Wu \textit{et al.}~\cite{mixed_darts} & Gradient-based & Poor & Any & No & None \\
EdMIPS~\cite{cai2020edmips} & Gradient-based & Poor & Any & No & None \\
FracBits~\cite{Yang2020FracBitsMP} & Gradient-based & Poor & Consecutive & Yes & None \\
Risso \textit{et al.}~\cite{risso2022igsc} & Gradient-based & High & Any & Yes & None \\ \hline
AutoQ~\cite{Lou2020AutoQ:} & RL & High & Any          & Yes  & Per-channel   \\
Gong \textit{et al.}\cite{fbnet_energy} & Gradient-based & High & Any          & No  & Per-channel, coarse-grain \\
Bayesian-bits~\cite{bayesian_bits} & Gradient-based & Poor & Power of two & No  & Per-channel   \\
DJPQ~\cite{diff_joint_pq} & Gradient-based & Poor & Any          & No  & {Per-channel}   \\
Chitty-Venkata \textit{et al.}~\cite{Chitty-Venkata2022} & Gradient-based & {High} & {Any}          & {No}  & Block-balanced (2:4)   \\
\hline
\textbf{Ours} & \textbf{Gradient-based} & \textbf{High} & \textbf{Any}          & \textbf{Yes} & \textbf{{Per-channel}}   \\ \hline

\end{tabular}
\begin{tablenotes}\footnotesize
\item [a] {High hardware-awareness indicates the usage of a cost objective tailored to the target hardware, as opposed to the number of ops. or parameters}
\end{tablenotes}
\end{threeparttable}
\end{adjustbox}
\vspace{-0.3cm}
\end{table*}

\subsubsection{Non-differentiable techniques}
\label{subsec:nondiff_techniques}
Several recent works have been devoted to the exploration of automatic methods to optimally assign different precisions to different parts of DNNs. 
The first attempts trying to solve the MPS problem considered only task performance as optimization goal. HAWQ-V2~\cite{hawq_v2} uses a sensitivity-based heuristic approach that considers second-order Hessian information. Lin \textit{et al.}~\cite{sqnr_mps} show an analytical method that optimizes signal-to-quantization-noise-ratio to find the optimal bit-width allocations across the layers of the network. 
HAQ~\cite{haq} and ReLeQ~\cite{releq} propose to use an iterative scheme, where an RL agent drives the bit-width assignment, at the granularity of individual layers, using both task performance and latency or energy measurements from the target hardware as rewards. 
Such approaches, while effective and flexible, suffer from the problem of long convergence times due to the iterative search space exploration, as discussed in Sec.~\ref{subsec:quantization}.

\subsubsection{Differentiable techniques}
One-shot algorithms emerged as a more lightweight solution with respect to the non-differentiable black-box optimization techniques. The earliest methods were based on a supernet scheme, inspired by the Differentiable NAS~(DNAS) literature~\cite{liu2018darts}. A multi-path network (the supernet) is constructed, which, in the case of MPS, contains all possible bit-width assignments as alternative paths. The optimization goal then reduces to selecting a single path from this network, and is solved through continuous relaxation using gradient-descent.
Wu \textit{et al.}~\cite{mixed_darts} propose one of the first approaches of this kind, where the 
optimization is performed considering a cost penalty term associated to the memory along with the standard task-specific loss.
Gong~\textit{et~al.}~\cite{fbnet_energy} propose a similar approach where precisions are explored on MobileNet-V2 architectures using as cost the energy consumption on the BitFusion accelerator. 

One of the key limitations of supernet-based MPS is the linear scaling of memory and computational complexity with the size of the search space. That is, adding a new path to the supernet with an operator quantized at a different precision requires duplicating the weights and computations associated with such an operator. For this reason, the search can be performed only on simpler and smaller proxy tasks, severely limiting the applicability of the method.
EdMIPS~\cite{cai2020edmips} solves this problem by substituting the expensive parallel convolutions of supernet-like approaches with an efficient composite convolution operation. In this approach, a single copy of the float tensors is fake-quantized on-the-fly at the different precisions explored (e.g., 2-, 4- and 8-bit) in each iteration of the gradient-based search.

All aforementioned works consider MPS on a per-layer basis, i.e., assigning precisions with the granularity of single layers, although independently for weights and activations.
To overcome this limitation, Yang \textit{et al.} propose FracBits~\cite{Yang2020FracBitsMP}, a gradient-based approach to perform \textit{kernel-wise} MPS. However, their method requires support for \textit{all bit-widths within a range}, and is not applicable in the common case of hardware platforms which support only some pre-defined precisions (e.g., MPIC~\cite{ottavi2020mpic}, that can operate with 2-, 4-, 8- and 16-bit data, but not, for instance with 3- or 5-bit).

Furthermore, most of the aforementioned methods optimize for a combination of task performance and hardware-agnostic complexity metrics such as network size and bitops count, i.e., the number of arithmetic operations performed multiplied by the precision of the operands~\cite{mixed_darts, bp-nas, cai2020edmips}. While size may effectively reflect the actual memory footprint of the network, metrics such as bitops correlate poorly with the real latency for executing a DNN on a target device~\cite{free_bits}. This is mainly due to the HW support for low-precision arithmetics, which in some situations may lead to a latency reduction not proportional to the bit-width, or even to an increase at lower precision~\cite{free_bits}.

Conversely, our previous work~\cite{risso2022igsc} considers the case of HW-aware differentiable channel-wise mixed-precision assignment for weights. Namely, it optimizes the precision of the weights of each convolutional layer channel independently, with a composite convolution operation inspired by EdMIPS~\cite{cai2020edmips}. Moreover, it uses a Look-Up Table (LUT)-based latency (or energy) model of the HW, in order to guide the search with a metric that well correlates with the actual performance on the hardware after the deployment.

\subsection{Joint Quantization and Pruning}
\subsubsection{Non-differentiable techniques}
The seminal work exploring the joint usage of pruning and quantization is Deep Compression~\cite{deep_compression}. Such approach is based on the cascaded application of first pruning and then quantization. While effective, applying the two methods separately may overlook mutual effects between the two. Moreover, Deep Compression does not consider MPS, but rather uses a fixed bit-width for the whole network.

APQ~\cite{Wang2020APQ} unifies NAS, structured pruning and MPS in a single pipeline by training a once-for-all network and an auxiliary neural network to predict the quantized accuracy of the sampled sub-networks. The search is conducted with an EA, and the MPS is performed layer-wise.
AutoQ~\cite{Lou2020AutoQ:} adopts an RL-based methodology to drive the bit-width selection for each kernel of a layer, considering also the option of pruning them away with a 0-bit assignment.
However, both APQ and AutoQ rely on iterative black-box techniques, thus suffering from long training times.
 
\subsubsection{Differentiable techniques}
Gong \textit{et al.}~\cite{fbnet_energy} propose a supernet-based differentiable approach that jointly quantizes and prunes bottleneck layers of MobileNet-V2. Namely, they build a supernet in which the alternatives represent different combinations of precision and pruning rate.
However, the explored search space is coarse with per-layer quantization, and pruning is achieved by only selecting the expansion factor of bottleneck layers.

DJPQ~\cite{diff_joint_pq} proposes a joint pruning and quantization scheme where quantization is non-linear and pruning is based on Gaussian gates that control the weight distribution of each layer's channel. They learn a variational posterior that tries to minimize the mutual information between successive layers' outputs. 
Bayesian-bits~\cite{bayesian_bits}
presents a holistic framework for structured channel pruning and mixed-precision quantization,
where pruning is considered as ``0-bit'' precision.
Each precision is associated to a learnable stochastic gate, and the objective function is a balance of task performance and network complexity, measured as bitops.
One limitation of this approach is that it is structurally able to explore only precisions which are power-of-two. This can be a drawback for HW platforms such as NE16, which support weights quantized to all possible bit-widths from 2 to 8. Note that this limitation is the mirror image of the one highlighted for FracBits~\cite{Yang2020FracBitsMP}.
Moreover, both DJPQ~\cite{diff_joint_pq} and Bayesian Bits~\cite{bayesian_bits} consider a per-layer mixed-precision scheme and rely on cost metrics that poorly correlate with real HW quantities, such as bitops~\cite{free_bits}.

Chitty-Venkata \textit{et al.}~\cite{Chitty-Venkata2022} propose a joint MPS and pruning approach based on gradient-descent. However, quantization is only performed layer-wise and pruning uses a 2:4 block-balanced approach with a fixed sparsity of 50\%. This limits the benefits of this approach to hardware platforms that support block-sparse operations.
Chitty-Venkata~\textit{et~al.}~\cite{Chitty-Venkata2023} similarly apply layer-wise MPS with a differentiable algorithm, proposing a method to eliminate suboptimal configurations from the search space.

Our method differs from all the aforementioned ones by combining gradient-based optimization with \textit{per-channel MPS} (plus joint pruning), and \textit{accurate HW models}. Moreover, our method does not impose any constraint on the candidate bit-widths, as opposed to Bayesian Bits~\cite{bayesian_bits} and FracBits~\cite{Yang2020FracBitsMP}.
A summary of the state-of-the-art MPS and pruning methods is reported in Table~\ref{tab:related}.

\section{Proposed Method} 
\label{sec:methods}
Mixed-precision quantization and pruning are orthogonal optimization techniques that allow to reduce the complexity of a DNN. 
However, they are usually applied sequentially, one after the other, especially when using a gradient-based optimization scheme. 
This sequential approach is suboptimal for two primary reasons: first, it leads to a higher training time required to obtain a final, optimized DNN. Second, the search space is limited as the second method's optimization options are constrained by the choices made during the first method's application, which is not reversible.

We thus propose a novel gradient-based method that performs both mixed-precision quantization and pruning at the same time. In this way, we can obtain a more deliberate precision assignment, considering also the option of pruning certain portions of the DNN in a structured form, instead of simply reducing their precision, if they do not positively impact the predictive performance. The search can be guided to explore the trade-off between predictive performance and DNN complexity, using cost models tailored specifically for the target hardware platform.

We remark that this paper only focuses on optimizing the precision assignment for a DNN in order to co-optimize its accuracy together with the efficiency (e.g. latency) on a given hardware target. The design of mixed-precision hardware for DNNs is outside the scope of this work.

In Sec.~\ref{subsec:search_space} we illustrate the designed search space; in Sec.~\ref{subsec:opt_method} we detail the optimization method, and in Sec.~\ref{subsec:regularizer} we explain the complexity regularizers that we define to drive the optimization towards low-cost solutions; in Sec.~\ref{subsec:train_proc} we describe the adopted training recipe and in Sec.~\ref{subsec:implementation_details} we discuss some implementation details related to the compatibility of our generated models with standard hardware and software for mixed-precision inference.

\begin{figure}[t]
  \centering
  \includegraphics[width=.49\textwidth]{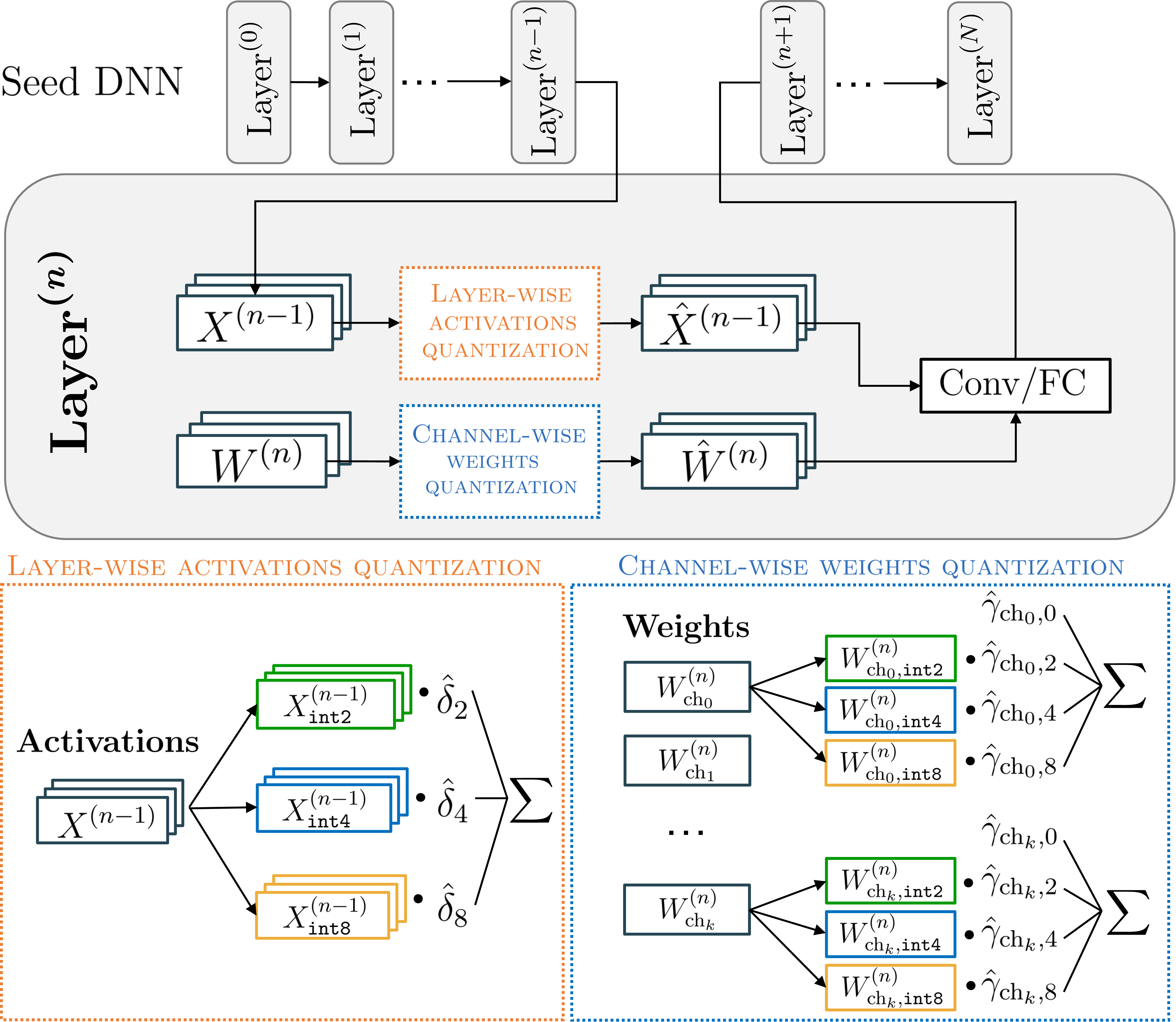}
  \vspace{-0.5cm}
  \caption{Overview of the quantization step of our proposed approach}
  \vspace{-0.4cm}
  \label{fig:figure_method}
\end{figure}

\subsection{Search Space} 
\label{subsec:search_space}
An overview of our method is depicted in Fig.~\ref{fig:figure_method}.
We define as $P_W$ and $P_X$ the sets of candidate precisions for the weights and activations, respectively. 
The search space comprises all the architectures that can be obtained from a reference DNN by quantizing its parameters and intermediate activations to all bit-widths in $P_W$ and $P_X$ respectively. 

As in our previous work~\cite{risso2022igsc}, we consider a \textit{channel-wise} precision assignment for weights, to have a finer search granularity.
Furthermore, we integrate pruning by considering an additional candidate precision, $p_0 \in P_W$, for the weights, which represents 0-bit quantization. Quantizing all weights of a given channel to 0 bits means setting them all to 0, effectively removing from the DNN's output all information conveyed by that channel. In fact, the channel's output activations will become constant. Thus, it is practically equivalent to pruning the channel away.

To define the search space, we associate each tensor of the neural network with a set of bit-width selection parameters. In particular, considering the $n$-th layer with $C^{(n)}_{\text{out}}$ output channels and its weights $W^{(n)}$, we define a matrix of bit-width selection parameters $\gamma^{(n)} \in \mathbb{R}^{C^{(n)}_{\text{out}} \times |P_{W}|}$, which associates to each output channel a vector of length equal to the cardinality of the weight precisions set, as shown in the bottom right part of Fig.~\ref{fig:figure_method}. 

The same procedure is applied to all the intermediate activations of the DNN, but in this case with layer-wise granularity. The rationale for supporting channel-wise quantization for weights but not activations is detailed in Sec.~\ref{subsec:implementation_details}. For each layer $n$, a vector of parameters $\delta^{(n)} \in \mathbb{R}^{|P_X|}$ is associated to the output activations, as depicted in the bottom-left part of Fig.~\ref{fig:figure_method}.
All bit-width selection parameters are optimized during the training phase, as explained more in detail in Sec. \ref{subsec:opt_method}.

Importantly, our method ensures that the same weights' channels are pruned away for pairs of layers in specific configurations, such as the two reconvergent layers of a residual block or a depthwise convolution following a pointwise convolution. 
This is obtained by sharing the precision selection parameters between those layers, and it is done to ensure that all pruned channels can be effectively removed from the network at the end of the optimization. For example, pruning away a certain channel of a residual branch's convolutional layer would result in lesser complexity reduction benefits if the corresponding channel in the other branch is not pruned as well, because their addition would require the channel to be kept in the following computations. If the channel is pruned from both branches, instead, one less feature map can be processed by the subsequent layers.
Similarly, each kernel of a pointwise convolution produces one feature map as output, which is then processed by a single depthwise kernel. Thus, if one of the output channels of the pointwise operation is pruned, also the associated depthwise filter can be removed, since it would process a "constant" feature map. 
In this example, only pruned channels shall be the same, whereas channels quantized at precisions greater than 0-bit could be assigned independently to the depthwise and pointwise layers. However, our method shares the entire bit-width selection parameters tensor between the two layers, to avoid over-complicating our method with multiple sets of masks. While this slightly restricts our search space, we found experimentally that it does not limit the effectiveness of our approach.

\subsection{Optimization Method}
\label{subsec:opt_method}
During the training phase, we optimize both the network's weights and the bit-width selection parameters simultaneously via gradient-descent. The training loss function includes two terms, to optimize respectively the performance of the network on the considered task and its complexity. In this way, the optimization favours the selection of lower precisions (down to 0-bit), in the least important portions of the network. The objective function is formulated as:
\begin{equation}
\label{eqn:loss_function}
    \min_{W, \theta} \left[ \mathcal{L}_{\text{task}}(W, \theta) + \lambda \mathcal{R}(\theta) \right]
\end{equation}
where $W$ is the set of network's weights, and $\theta := \{\delta, \gamma\}$ is the set of bit-width selection parameters. The task loss term $\mathcal{L}_{\text{task}}(W, \theta)$ depends on both $W$ and $\theta$. On the other hand, the regularization loss term $\mathcal{R}(\theta)$ is only affected by the bit-width selection parameters, and must be designed to model the desired cost metric, such as size or latency, in a differentiable manner. The relative balance between the two terms is controlled by the scalar strength hyper-parameter~$\lambda$. A higher value of $\lambda$ biases the search process towards finding more lightweight networks.

In the search phase, for each weight channel or activation tensor of supported layers (convolutional and linear), an effective tensor is computed, as represented in Fig.~\ref{fig:figure_method}. 
More in detail, considering an activation tensor $X^{(n)}$ and its associated bit-width selection parameters $\delta^{(n)}$, the first step consists in obtaining a discrete probability distribution from the latter. Thus, we obtain a new vector $\hat{\delta}^{(n)}$ whose elements are constrained in the $[0,1]$ range and sum up to 1. This can be achieved by applying one among different sampling methods. In this paper we consider three sampling methods, namely a softmax operation (SM), an argmax (AM) operation and a hard Gumbel-Softmax operation (HGSM). 
We can thus define the sampling operation for the bit-width selection parameters as $h(\delta)$, such that for $i =0, ..., |P_X|$:
\begin{equation}
\label{eqn:sampling_functions}
    \hat{\delta}_i = h(\delta)_i =
    \begin{cases}
      \frac{\textstyle\exp(\delta_i / \tau)}{\textstyle\sum_j \exp(\delta_j / \tau)}, & \text{SM}\\[10pt]
      \frac{\textstyle\exp(\delta_i / \tau)}{\textstyle\sum_j \exp(\delta_j / \tau)}, \tau \to 0, & \text{AM} \\[10pt]
      \frac{\textstyle\exp\left[(\delta_i + \epsilon) / \tau\right]}{\textstyle\sum_j \exp\left[(\delta_j + \epsilon_i) / \tau\right]}, \tau \to 0, & \text{HGSM}
    \end{cases}
  \end{equation}
where $\tau$ is a temperature hyper-parameter that controls the smoothness of the distribution and $\epsilon_i \sim \text{Gumbel}(0,1)$ is an i.i.d. sample drawn from the Gumbel distribution. 
Then, the effective tensor $\hat{X}^{(n)}$ is computed as:
\begin{equation}
    \hat{X}^{(n)} = \sum_{p_x \in P_X} \hat{\delta}_{p_x}^{(n)} \cdot X_{p_x}^{(n)}
\end{equation}
where $X_{p_x}$ is the activation tensor quantized at $p_x$ bits.
The effective activation tensor is thus a linear combination of its variants quantized at all the candidate precisions $p_x \in P_X$. 

A similar approach is applied to the weights of the DNN. Each $k$-th row of the matrix $\gamma^{(n)}$ contains a vector of bit-width selection parameters for the $k$-th channel of the $n$-th layer. To obtain a probability distribution for each of the channels, we apply one of the aforementioned sampling functions to each row $\gamma^{(n)}_k$, as in Eq.~\ref{eqn:sampling_functions}, obtaining $\hat{\gamma}^{(n)}_k = h\left(\gamma^{(n)}_k\right)$.
We then compute the effective weights as:
\begin{equation}
\label{eqn:effective_weight}
    \hat{W}^{(n)} = \sum_{p_w \in P_W} \hat{\gamma}_{p_w}^{(n)} \cdot W_{p_w}^{(n)}
\end{equation}
with $W_{p_w}$ being the weight matrix quantized at $p_w$ bits. This time, $\hat{\gamma}_{p_w}^{(n)}$ is a vector with $C_{\text{out}}$ elements which contains the bit-width selection parameters associated to the $p_w$ bit-width for all the output channels.

The effective activations and weights tensors are used to produce the output $Y^{(n)}$ of the $n$-th layer:
\begin{equation}
    Y^{(n)} = l(\hat{X}^{(n-1)}, \hat{W}^{(n)})
\end{equation}
where $l$ represents the operation performed by the layer.

At the end of the training process, one single precision bit-width must be assigned to each weights' channel and activation. Each tensor is thus replaced by its quantized version at the precision corresponding to the highest value in the bit-width selection parameters vector. Namely, for the $n$-th layer's activation:
\begin{equation}
    X^{(n)} \leftarrow X^{(n)}_{p_x}\ |\ p_x := \underset{p_x \in P_X}{\text{argmax}}(\hat{\delta}^{(n)})
\end{equation}
Whereas for weights, each $k$-th channel can be quantized at a different precision:
\begin{equation}
     W^{(n)}_k \leftarrow W^{(n)}_{k, p_w}\ |\ p_w := \underset{p_w \in P_W}{\text{argmax}}(\hat{\gamma}_k^{(n)}) 
\end{equation}

Notably, before applying the search procedure described in this section, we implement batch normalization folding with the preceding convolutional or linear layer. 
The rationale is that hardware equipped with solely integer arithmetic units would not support floating point batch normalization parameters. Thus, by emulating a full-integer inference already in the optimization phase, we match more closely the final deployed networks.

\subsection{Complexity Regularizers}
\label{subsec:regularizer}
The methodology detailed in Sec.~\ref{subsec:search_space} and Sec.~\ref{subsec:opt_method} is orthogonal to the cost metric selected as proxy for the DNN's complexity ($\mathcal{R}$ in Eq.~\ref{eqn:loss_function}). Thus, any measure of DNN cost can be applied to guide the training process, as long as it is expressed by a differentiable function, given that it must be included in a gradient-based optimization loop.
Commonly used cost functions are hardware-agnostic proxies for general properties of a DNN, like its size or its number of Multiply-and-Accumulate (MAC) operations~\cite{chitty-venkata_neural_2023}. 
While model size is a good proxy for memory occupation, the number of MACs often correlates poorly with the measured performance (e.g., inference latency or energy consumption) on the real hardware~\cite{cai_enable_2022}. Therefore, more accurate regularizers can be introduced, which take into account a given platform's support for each combination of weights and activation precision, and the corresponding efficiency.

In our work, we mainly focus on model size as hardware-agnostic cost metric. Moreover, we introduce two hardware-specific regularizers, to estimate the inference latency on two deployment targets, namely the Mixed Precision Inference Core (MPIC)~\cite{ottavi2020mpic} and the Neural Engine 16 (NE16)~\cite{macan2023ne16}, as detailed below.

\subsubsection{Size Regularizer}
If the optimization objective is to minimize the amount of memory required to store the model's parameters, the activations' bit-widths do not have any impact, since they only influence the run-time memory occupation. Thus, the corresponding regularizer considers the effective weights bit-width, that is,  the dot product between the precision bit-widths and the corresponding selection parameters vector for each channel $i$ of the $n$-th layer. The regularizer for a convolutional layer can then be written as:
\begin{equation}
\label{eqn:size_reg}
    \mathcal{R}^{(n)}(\hat{\gamma}^{(n)}) = C_{\text{in,eff}}^{(n)} K_x^{(n)} K_y^{(n)} \sum_{i=1}^{C_{\text{out}}^{(n)}} \sum_{p_w \in P_W} \hat{\gamma}^{(n)}_{i, p_w} p_w
\end{equation}
where $K_x^{(n)}$ and $K_y^{(n)}$ are the horizontal and the vertical kernel size. The term $C_{\text{in,eff}}^{(n)}$ represents the effective number of input channels, i.e., those which have not been pruned away. 
Specifically,  for a simple case of sequentially-connected layers, $C_{\text{in,eff}}^{(n)}$ is computed as $C_{\text{out}}^{(n-1)} - \sum_{i=1}^{C_{\text{out}}^{(n-1)}} \hat{\gamma}_{i, p_0}$.
Using $C_{\text{in,eff}}$ instead of $C_{\text{in}}$ in the cost function models the fact that pruning an output feature map also provides benefits to the subsequent layers since they will process a smaller input.

\subsubsection{Mixed Precision Inference Core Regularizer}
\label{sec:mpic}
The Mixed Precision Inference Core~(MPIC)~\cite{ottavi2020mpic} is a RISC-V core based on the open-source RI5CY, featuring a 4-stage pipeline and supporting all standard RISC-V extensions, together with the XpulpV2, a DSP-oriented extension that includes hardware loops, post-increment loads/stores, and single-instruction multiple-data (SIMD) instructions. 
MPIC extends this core by including the new XMPI extension, which adds SIMD support down to 2 bits.
Specifically, MPIC includes a dedicated dot-product unit to parallelize MAC operations on inputs (weights and activations) independently quantized. It can perform 2$\times$16-bit, 4$\times$8-bit, 8$\times$4-bit, or 16$\times$2-bit operations in parallel, thus gaining speedup from lower bit-widths. For mixed-precision MACs, the smallest operand is sign-extended through dedicated HW. An additional speedup is anyway achieved compared to the homogeneous-precision, thanks to the reduced amount of fetch operations from memory.

To quantify the latency cost of executing a DNN on MPIC, we use the Look-Up Table from~\cite{ottavi2020mpic}, which comprises latency measurements collected on fixed layer topologies. In particular, the LUT stores the number of MAC operations per cycle for all combinations of activations and weights bit-widths.
For a single convolutional layer $n$, the MPIC regularizer $\mathcal{R}^{(n)}(\hat{\delta}^{(n-1)}, \hat{\gamma}^{(n)})$ is defined as: 
\begin{equation}
\label{eqn:mpic_regularizer}
    \sum_{p_x \in P_X}  \sum_{p_w \in P_W, p_w \neq 0} \frac{ \mathcal{M}^{(n)}( \hat{\delta}^{(n-1)}_{p_x}, \hat{\gamma}^{(n)}_{p_w})} {\mathcal{T}(p_x, p_w)} 
\end{equation}
where $\mathcal{M}^{(n)}$ represents the MACs operations executed at the different $p_x/p_w$ combinations in the $n$-th layer and $\mathcal{T}$ represents the LUT. The number of MACs ($\mathcal{M}^{(n)} ( \hat{\delta}^{(n-1)}_{p_x}, \hat{\gamma}^{(n)}_{p_w})$) is computed as:
\begin{equation}
    K_x^{(n)} K_y^{(n)} W^{(n)} H^{(n)} C^{(n)}_{\text{in,eff}}\ \hat{\delta}^{(n-1)}_{p_x}\  \sum_{i=1}^{C_{\text{out}}}  \hat{\gamma}^{(n)}_{i,p_w}
\end{equation}
where $\sum_{i=1}^{C_{\text{out}}}  \hat{\gamma}^{(n)}_{i,p_w}$ represents the theoretical number of output channel executed at the target $p_w$ precision with $\hat{\gamma}^{(n)}_{i,p_w}$ the quantization parameter of the $i$-th channel for the $p_w$ weights' bit-width, $C^{(n)}_{\text{in,eff}}\ \hat{\delta}^{(n-1)}_{p_x}$ represents the theoretical portion of input activations processed at the target $p_x$ bit-width and where $W^{(n)}$ and $H^{(n)}$ are, respectively, the width and height of the $n$-th layer's output.
In essence, Eq.~\ref{eqn:mpic_regularizer} computes the number of MACs executed at every combination of $p_x$/$p_w$ precisions in the layer and then uses the MACs/cycle value in the LUT to determine the cost as a number of cycles.
For the results reported in Table~\ref{tab:deployment_results}, we compute the latency with Eq.~\ref{eqn:mpic_regularizer} and the energy consumption associated to an individual inference from the power measurements reported in\cite{ottavi2020mpic} collected with the core running at a frequency of 250 MHz.

\subsubsection{Neural Engine 16 Regularizer}
\label{sec:ne16}
Neural Engine 16 (NE16)~\cite{macan2023ne16} is an instance of the NEeureka accelerator~\cite{prasad_specialization_2023} composed of 3$\times$3 processing elements (PE) included in the commercial GAP9 System-on-Chip\footnote{\texttt{\url{https://greenwaves-technologies.com/gap9_processor/}}}.
Each PE is responsible for computing an individual convolution output spatial pixel over 32 output channels.
The accelerator is tailored to perform $3\times3$ convolutions, $1\times1$ pointwise, and $3\times3$ depthwise operations, and supports 8-bit quantization for activation tensors and 2- to 8-bit for weights.
As basic blocks, NE16 includes 1×8-bit multiplier arrays (i.e., parallel AND gates), which are used in different configurations depending on the selected operating mode.
Specifically, for $p_w > 1$, the weight bits are distributed to multiple basic blocks and the outputs are then combined appropriately. Accordingly, the latency of MAC operations increases proportionally to the weights precision.

The cost model that we use to determine the number of execution cycles given the precision assignments takes into account three main factors.
(i) The weights loading latency, which depends on the data STREAMER - a custom memory access
engine, which can reach up to 288 bits of bandwidth; 
(ii) The time required by the PE matrix to complete its MAC operations, which depends on the number of input channels, the spatial dimensions, and the output channels, taking into account that each PE element can execute up to 3$\times$3$\times$32 parallel MAC operations, depending on the operating mode;
(iii) The latency to store the results in the L1 memory, which is given by the normal bandwidth of GAP9, i.e., 64 bits per cycle.
For the sake of space, we do not report the complete analytical model, but we refer readers to its open-source implementation\footnote{\texttt{\url{https://github.com/pulp-platform/dory/blob/master/dory/Hardware_targets/PULP/GAP9_NE16/Tiler/Ne16PerfModel.py}}}.

We compute the total latency of our DNNs considering that the accelerator runs at the GAP9's maximum operating frequency of 370 MHz. We do not report the energy estimation for single inferences because no power measurements are publicly available for this DNN accelerator.

On this platform, we also implement a post-search refinement step to our precision assignment, aimed at fully exploiting the parallelism of the accelerator. In fact, the assignment obtained from the gradient-based search can sometimes have a mismatch with the hardware parallelism (e.g. 33 channels assigned to a given precision, requiring a second NE16 invocation just for 1 additional channel). In those cases, we use a deterministic optimization step to consider \textit{increasing} (but never decreasing) the bit-width of some channels, if that leads to a reduction in inference latency. This post-processing does not require additional training, and takes less than 1~s on our models. Moreover, note that it can be applied not only to NE16, but to any target which parallelizes processing over output channels.

\subsection{Training Procedure} 
\label{subsec:train_proc}
The procedure to apply the proposed joint mixed-precision quantization and pruning optimizations comprises three distinct phases: warmup, search and fine-tuning. The warmup phase consists in the training of the DNN's weights only, without the additional bit-width selection parameters. Weights are trained in floating point, and optimized only with respect to the task loss, whereas the regularization loss is not applied. The DNN generated by this preliminary training serves as starting point for the subsequent optimization phases. 

The search phase represents the core of the optimization. The bit-width selection parameters are trained together with the DNN's weights. In this step, the complexity-dependent regularization component ($\mathcal{R}$) is applied to the loss. 
Throughout the search phase, when the softmax sampling is selected, the temperature $\tau$ is gradually lowered in order to eventually make the sampling of the bit-width selection parameters resemble an argmax operation. This is useful when the distribution of the sampled bit-width selection parameters is spread-out, but the major contribution is associated to the 0-bit precision. During training, the layer would get used to a non-zero output, given the contributions from all the bit-widths, but after the final discretization, the channel would be pruned, yielding inconsistent behaviours with respect to the training phase.

At the end of the search phase, the bit-width selection parameters are discretized as explained in Sec.~\ref{subsec:opt_method} in order to assign  a single precision to each weight channel and intermediate tensor. Then the fine-tuning phase trains the final quantized version of the DNN until convergence, optimizing only its weights according to the task loss term $\mathcal{L}_{\text{task}}$.

\subsubsection{Weights rescaling}
As shown in Eq.~\ref{eqn:effective_weight}, the effective weight tensor is a weighted average of the quantized versions of the original weight matrix. 
However, differently from other quantizations which all approximate the floating point weights at different precisions, 0-bit quantization always produces a constant zero output. If the corresponding bit-width selection parameter is non-null at the beginning of the search phase, this will lower the magnitude of the effective weight tensor with respect to the post-warmup value, causing a significant accuracy drop, which in turn will lead to suboptimal precision assignments.
We counteract this problem by rescaling the weights of the post-warmup model in such a way that the 0-bit quantization does not systematically lower the effective weight tensor. In particular, weight channels at the beginning of the search phase are assigned as follows:
\begin{equation}
    W_i^{(n)} \leftarrow \frac{W_i^{(n)}}{\underset{p_w \in P_W, p_w \neq 0}{\sum} \hat{\gamma}^{(n)}_{i, p_w}}
\end{equation}

\subsubsection{Bit-width selection parameters initialization}
Precision selection parameters for weights and activations ($\gamma$ and $\delta$) are initialized based on the assumption that low bit-widths and pruning should be favoured in later epochs, to avoid an excessive performance drop at the beginning of the search phase. Furthermore, initializing all $\gamma$ values uniformly would lead to instabilities when using a discrete sampling method, since the latter might lead to the choice of the 0-bit precision for a large portion of the network, or even an entire layer, causing an interruption in the backpropagation of the gradients. Thus, we initialize the weights' bit-width selection parameters as follows:
\begin{equation}
    \hat{\gamma}_{i, p_w}^{(n)} \leftarrow \frac{p_w}{\underset{p \in P_W}{\max} p},\ \forall p_w \in P_W
\end{equation}
In essence, we assign progressively decreasing values to lower bit-widths, which consequently leads to reduced associated sampling coefficients, with the lowest value associated to 0-bit (i.e., pruning).
We perform an analogous assignment also for $\delta$ parameters.
In this way we ensure that, in the first training steps, the highest precisions are selected with much higher probability, avoiding instabilities. 

\begin{figure}[]
  \centering
  \includegraphics[width=.49\textwidth]{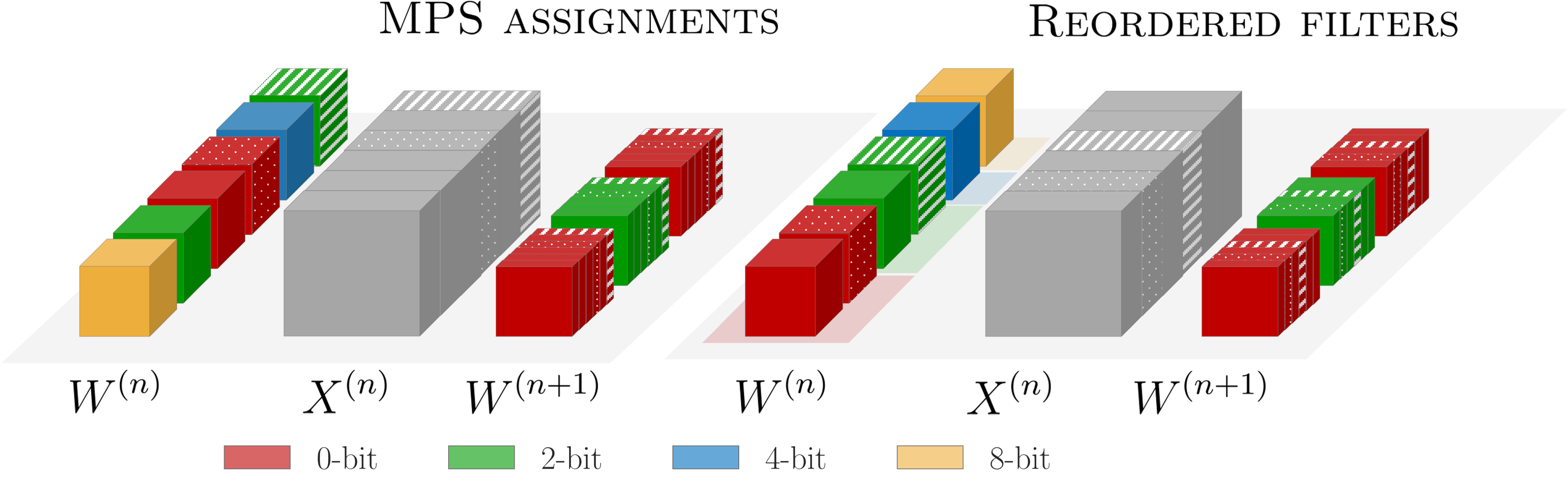}
  \vspace{-0.7cm}
  \caption{Reordering of the weights channels by bit-width after the precision assignment}
  \vspace{-0.3cm}
  \label{fig:filters_reordering}
\end{figure}

\subsection{Implementation details}
\label{subsec:implementation_details}
Our proposed approach, which assigns different precisions to the channels of each layer of a DNN, is fully compatible with hardware supporting mixed-precision inference and with software libraries. As shown in the leftmost part of Fig.~\ref{fig:filters_reordering}, after the search phase, each channel of the weight tensor of a given DNN's layer can be quantized at a different precision, with no specific order. However, to allow an efficient execution on an edge device, we can split and reorganize the channels in different groups according to the bit-width at which the weights are quantized, as depicted in rightmost part of the figure. This transformation is performed only once, offline, before the inference. The change in the weight channels' order influences also the produced output activations, as shown by the matching patterns in Fig.~\ref{fig:filters_reordering}. In turn, also the subsequent layers' weights must be reordered accordingly, in order to maintain the correct association between input channels and weights. 

After this reordering process, the original convolutional (or linear) layer can be split into a set of $|P_W|$ distinct and concurrent smaller sub-layers, as depicted with the coloured shadows in Fig.~\ref{fig:filters_reordering}. Since the activations are instead quantized layer-wise, the splitted convolutions' outputs all have the same precision and can be easily concatenated, before being fed to the subsequent layer. Having the activations quantized with a channel-wise scheme as well, would instead lead to a mixed-precision output, much more difficult to store in memory so that it can be read efficiently by subsequent layers. For this reason, we leave the study of channel-wise activation quantization to future work.

Importantly, our proposed channel-wise MPS implements weight sharing, i.e. all the quantized weights that are combined as illustrated in Eq.~\ref{eqn:effective_weight} originate from the same real-valued weight tensor. Thus, by incrementing the number of candidate precisions, there is only a negligible memory overhead associated to the definition of one additional quantization parameter per channel, rather than an entire copy of the weights. This overhead is small regardless of the size of the network. Moreover, given the low number of additional parameters to be optimized, the time overhead of the training process is also comparable with that of layer-wise MPS for the same candidate precisions set.

\section{Experimental Results}
\label{sec:results}
\begin{figure*}[]
  \centering
  \includegraphics[width=.96\textwidth]{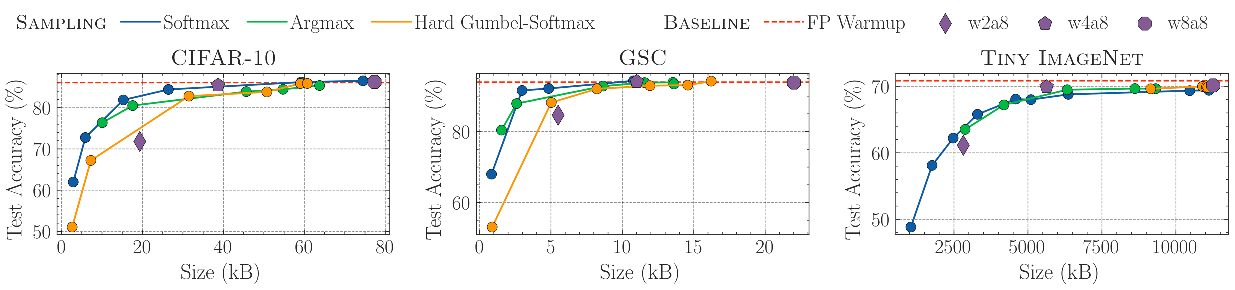}
  \vspace{-0.45cm}
  \caption{Results obtained with our proposed approach on CIFAR-10, GSC and Tiny ImageNet. Different sampling methods are compared.}
  \vspace{-0.3cm}
  \label{fig:pareto_analysis_right}
\end{figure*}
\begin{figure*}[t]
  \centering
  \includegraphics[width=.96\textwidth]{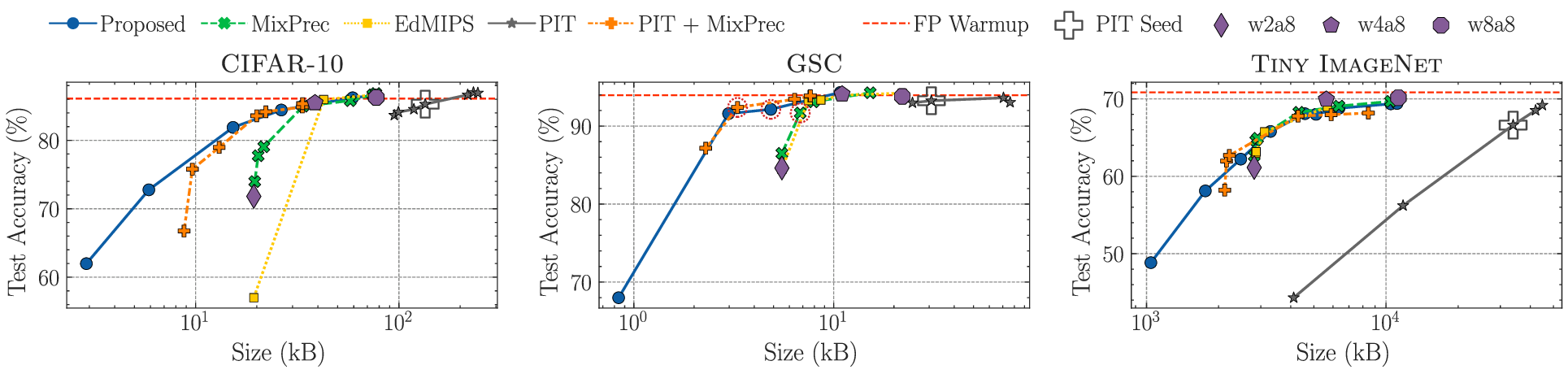}
  \vspace{-0.45cm}
  \caption{Comparison of the results obtained by our proposed method with other state-of-the-art approaches. The architecture optimized by PIT that is used as input for MixPrec is denoted as \emph{PIT Seed}.}
  \vspace{-0.35cm}
  \label{fig:sota_comparison}
\end{figure*}

\subsection{Experimental Setup}
\label{sec:experimental_setup}
We implement our method in Python 3.10.9 and PyTorch 1.13.1, extending the PLiNIO library~\cite{plinio_fdl}.
We consider as candidate precisions 0, 2, 4 and 8 bits for the weights of the DNNs, since they are well supported by all considered target accelerators. We use different activations' bit-width sets for different experiments.
We use the PACT~\cite{choi2018pact} quantization scheme for the activations and a symmetric min-max quantization strategy for the weights.

We compare our method with several state-of-the-art approaches. In particular, we consider an extension of the PIT mask-based DNAS~\cite{risso2023pit}, targeting also 2D convolutions. PIT implements channel pruning with a masking mechanism similar to our proposed method. Moreover, we also consider EdMIPS~\cite{cai2020edmips}, a state-of-the-art gradient-based mixed precision quantization method, and the channel-wise MPS proposed by Risso~\textit{et~al.}~\cite{risso2022igsc} (denoted MixPrec in the following Sections). Lastly, we compare against the sequential application of PIT and MixPrec, to assess the goodness of our method against the usual flow of applying pruning and quantization sequentially. As mentioned in Sec.~\ref{subsec:regularizer}, we deploy our mixed-precision networks on MPIC and NE16.

We run experiments on three different benchmarks, namely CIFAR-10\footnote{\texttt{\url{http://www.cs.toronto.edu/~kriz/cifar.html}}}, Google Speech Commands (GSC) v2~\cite{google-speech-commands} and Tiny ImageNet\footnote{Le, Y., \& Yang, X.S. (2015). Tiny ImageNet Visual Recognition Challenge.}.
For CIFAR-10 we divide the images into training, validation and test sets in the proportions of 66\%, 17\% and 17\%, respectively. The reference architecture is a custom ResNet with 9 convolutional layers, as in~\cite{mlperftiny-benchmark}. 
For GSC, as in~\cite{google-speech-commands}, we consider the standard classification setting comprising only 12 target labels, i.e., 10 keywords from the original dataset and two additional "Silence" and "Unknown word" classes. The training, validation and test sets include, respectively, 85\%, 10\% and 5\% of the data samples. The reference architecture is the Depthwise Separable Convolutional Neural Network (DS-CNN) from~\cite{mlperftiny-benchmark}.
For Tiny ImageNet, we divide the original training set into a training and a validation partition with a 90:10 split. Then, we used the official validation partition as test set. The baseline DNN is a ResNet-18. All accuracy results refer to the respective test sets.

\subsubsection{Training protocol} 
We set the training epochs for the warmup, search and fine-tuning phases to 500, 200 and 50 for the CIFAR-10, GSC and Tiny~ImageNet benchmarks, respectively. 
The cross-entropy loss is used as task loss $\mathcal{L}_{\text{task}}$. For the GSC benchmark, due to the imbalance of the dataset, we employ class weights equal to the inverse of the frequency of the classes' samples.

We use the SGD optimizer with a learning rate of 1e-2 and momentum of 0.9 to optimize the bit-width selection parameters on all the three considered tasks. For CIFAR-10 and GSC, we use the Adam optimizer with learning rate equal to 1e-3 and weight decay of 1e-4 for the DNN's weights; while for Tiny~ImageNet we use the SGD optimizer with learning rate equal to 5e-4, momentum of 0.9 and weight decay of 1e-4. For CIFAR-10, for both the weights' and bit-width selection parameters' optimizers, we reduce at each epoch the learning rate by a factor of $0.99$; while we decay the learning rate by 0.1 every 7 epochs for Tiny~ImageNet. For GSC we halve the learning rate at epochs 50 and 100, and we divide it by 2.5 at epoch 150.

We apply early stopping with patience equal to 50, using as control metric the validation accuracy for CIFAR-10 and Tiny~ImageNet, and the validation loss for the GSC benchmark due to the dataset's imbalance. 

The initial softmax temperature is set to 1, and lowered at every epoch by $e^{-0.045}$ for CIFAR-10 and GSC as proposed in~\cite{wan2020fbnetv2}, and by 0.638 for Tiny~ImageNet. The rationale behind the latter value is to obtain the same final temperature despite the fewer training epochs.
To ensure a fair comparison also in terms of training budget between our method and the reference models quantized at fixed-precision, we train all baselines for a total number of epochs equal to the sum of warmup, search and fine-tuning epochs.

\subsection{Sampling Methods Comparison}
\label{sec:results_pareto_analysis}
Fig.~\ref{fig:pareto_analysis_right} reports the results obtained by applying our method on the three benchmarks, using the size regularizer as cost metric. Each point in the curves represents a model trained with a different value of the regularization strength ($\lambda$). Only the DNNs belonging to the Pareto front according to the validation metrics are reported.
Each plot shows as baselines the reference architecture trained in floating point (FP) and three fixed-precision versions with the weights quantized at 2, 4, and 8 bits. 
The activations are quantized at 8 bits since they do not impact the model size. 
Each color of the curve is associated to a different sampling method. 

The leftmost plot shows the results on CIFAR-10. We obtain three rich Pareto fronts of architectures, with varying size between few kilobytes up to 77.12~kB. It is noticeable how the softmax sampling consistently outperforms the other methods. 
At iso-accuracy with respect to the floating point seed model, which requires 309.44~kB for the parameters, we achieve a 80.86\% reduction in model size; and a reduction of 23.43\% with respect to the fixed-precision w8a8 baseline. 
Furthermore, our proposed method is able to push the model size reduction even further at a cost of few percent in accuracy. 
With an accuracy drop of only 1.67\%, we have a model of size equal to only 26.41~kB, i.e., 91.46\% smaller than the floating point seed model.
At iso-accuracy with the w2a8 reference model, i.e. 71.81\%, we obtain a model of only 5.89~kB in size, 98.10\% and 69.54\% less than the seed and the w2a8 models, respectively.

The middle plot reports the results on GSC. Also in this case, the softmax sampling method yields the best trade-offs in the accuracy vs. model size space. With an increase in accuracy equal to 0.37\%, we obtain an 87.76\% decrease in the model size with respect to the floating point seed, whose size is 88.06~kB.
At iso-accuracy with respect to the 8-bit baseline, we obtain a solution associated to a 47.50\% smaller size.
Furthermore, with a reduction equal to 2.35\% in accuracy with respect to the FP seed, we have a model of 2.98~kB in size, 96.62\% and 45.90\% smaller than the seed and the w2a8 baselines, respectively, while also being more accurate (+6.97\%) than the latter.

On Tiny ImageNet (rightmost plot), the softmax sampling leads to the best trade-offs in the smallest sizes region. For higher sizes, instead, the argmax and hard Gumbel-Softmax methods yield slightly better results in terms of predictive performance. However, the differences are mostly negligible.
On this more complex benchmark we do not obtain any model at iso-accuracy with respect to the FP seed, although we approach it closely. The most accurate model achieves 70.08\% in accuracy, with a 75.48\% size reduction with respect to the FP seed (45.05~MB), and a 1.91\% size reduction compared to the w8a8 baseline, while being slightly less accurate (-0.06\%) than the latter.
While the method does not lead to any solutions that dominate the w4a8 reference model, it yields significant enhancements for higher regularization strengths. At iso-size with respect to the w2a8, we improve accuracy by 2.39\%. Furthermore, with an additional 1.08\% of accuracy with respect to w2a8, we have a 12.15\% smaller size.

From these experiments, we conclude that Softmax is the most stable sampling strategy, since it yields the best results on CIFAR-10 and GSC, and has a comparable performance with the other sampling methods on Tiny ImageNet. Thus, we employ Softmax sampling in all following Sections.

\subsection{State-of-the-art Comparison}
\label{sec:results_sota_comparison}
Fig.~\ref{fig:sota_comparison} compares our method against state-of-the-art approaches.
The selection of the architectures belonging to the Pareto curves is based on the validation accuracy vs. model size trade-off. To evaluate the concatenation of PIT and MixPrec, we first obtain a set of architectures by applying PIT to the floating point model, then choose one of them as seed for MixPrec (shown with a black plus symbol). Lastly, we use this seed to perform a new set of MixPrec searches (by varying the $\lambda$ in Eq.~\ref{eqn:loss_function}) to obtain the final Pareto front.

In the leftmost plot of Fig.~\ref{fig:sota_comparison} we report the results on CIFAR-10. In the size range delimited by the w4a8 and w8a8 baselines, the results obtained by our method are comparable to the ones achieved by all other MPS approaches. The differences reveal themselves for smaller sizes, i.e., higher regularization strengths, where the curves exhibit distinct trends. In particular, EdMIPS and MixPrec have a tight lower bound in terms of size, associated to the assignment of the lowest bit-width to the entire DNN. This leads to the same size value obtained by the fixed-precision w2a8 baseline. The associated accuracy instead varies due to the different weights' optimizations carried out in the search phase. In particular, with EdMIPS we obtain a large accuracy drop at iso-size with the w2a8 baseline because the training could not converge properly.
Our method can instead overcome such lower bound and find Pareto solutions with lower sizes, thanks to the application of pruning. When compared to the results obtained with the sequential application of PIT and MixPrec, our optimization method yields comparable or superior trade-offs. This is due to the fact that we allow a broader flexibility in the precision assignments in the DNN. Instead, when applying PIT first, and MixPrec in a subsequent step, the former narrows the latter's search space, potentially leading to suboptimal solutions.

We present the results on GSC in the central plot of Fig.~\ref{fig:sota_comparison}. EdMIPS and MixPrec yield comparable results with each other, with very similar Pareto fronts.
Our method achieves an accuracy of 91.60\% with a model size of 2.98~kB. At iso-accuracy, we reduce size by 56.17\% with respect to the 6.79~kB model found by MixPrec.
Our method also leads to comparable results with respect to the concatenation of PIT and MixPrec. This is due to the fact that the reference FP model is highly over-sized, thus the pruning performed by PIT does not remove fundamental channels which cannot be recovered afterwards. Furthermore, quantization at low bit-widths does not wreck accuracy, allowing MixPrec to reduce the precision with negligible performance impact.

However, as shown in Table~\ref{tab:avg_speedup}, our method significantly reduces the total search time with respect to the concatenation of PIT and MixPrec.
Individually, a single PIT epoch requires $1.8 \times$ more time than a standard training epoch of the reference DNN, while for both MixPrec and our proposed approach, the overhead is $4.3 \times$. On the other hand, obtaining a seed for MixPrec when the two methods are applied sequentially requires to first derive the full Pareto curves of architectures with PIT. Thus the overhead to obtain a single solution is $(1.8 N + 4.3)$ times the floating point model's training time, where $N$ is the total number of models trained with PIT, which can be higher than the number of Pareto points. 
For instance, for GSC, we had to train 4 PIT models to obtain the Pareto front shown in Fig.~\ref{fig:icl_mpic_ne16}, resulting in a total overhead of $11.5\times$. This is approximately $2.7\times$ higher than the total time required by our joint method to obtain the same result.

In the rightmost plot of Fig.~\ref{fig:sota_comparison}, we report the results on Tiny~ImageNet. As observed for the other benchmarks, the different methods lead to similar results for larger DNNs. The benefits of our approach are most visible for smaller models. 
With a model 12.15\% smaller than the w2a8 baseline, we achieve a higher accuracy (62.21\% vs. 61.13\%). The concatenation of PIT and MixPrec yields slightly better results, with a 10.29\% size reduction at iso-accuracy with respect to our solution.
With an accuracy of 58.11\%, our proposed method is instead able to achieve a 16.96\% size reduction with respect to PIT + MixPrec at iso-accuracy. Furthermore, we have a $\approx3.1 \times$ lower search time overhead (considering the 5 models on the PIT Pareto front).

\subsection{Deployment}
\label{sec:results_deployment}
\begin{figure}[t]
  \centering
  \includegraphics[width=\columnwidth]{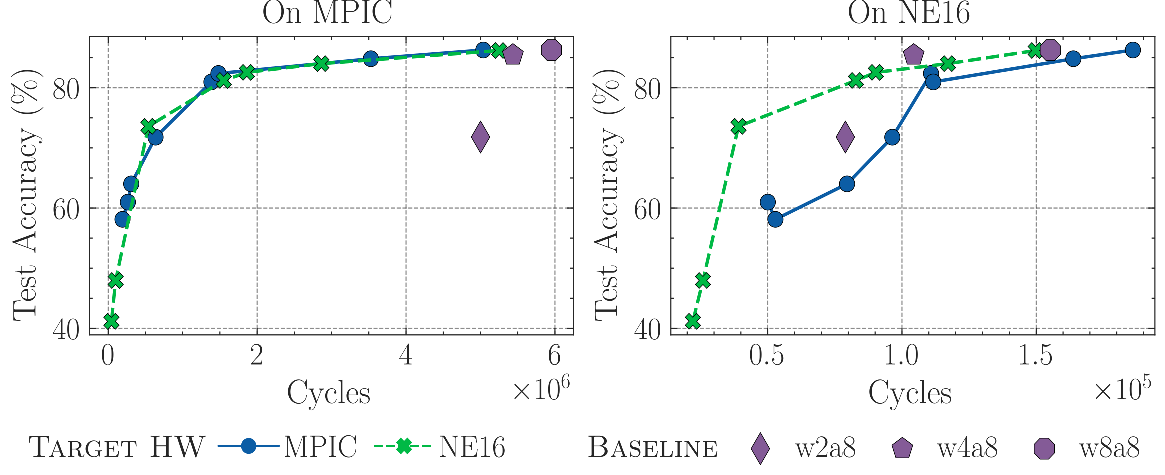}
  \vspace{-0.8cm}
  \caption{Comparison in the accuracy vs. cycles space of models trained with different cost regularizers, on CIFAR-10. The leftmost plot presents the results obtained when considering MPIC as actual deployment target. The rightmost plot those with NE16 as deployment target.}
  \label{fig:icl_mpic_ne16}
  \vspace{-0.4cm}
\end{figure}

In Fig.~\ref{fig:icl_mpic_ne16} we report the results obtained on CIFAR-10 by coupling our method with the latency cost models for MPIC and NE16 described in Sec.~\ref{subsec:regularizer}. Also in this experiment, we kept the activations at 8 bits to have a fair comparison between the models obtained with the two cost models, since it is the only bit-width that NE16 supports for the intermediate tensors. The green dashed curves represent the Pareto fronts obtained by applying the NE16 latency cost model explained in Sec.~\ref{sec:ne16}. The blue curves instead are obtained by applying the MPIC cost model of Sec.~\ref{sec:mpic}. 

\begin{table}[t]
\centering
\caption{Average total training time speed-up of our approach with respect to the sequential application of PIT and MixPrec}
\label{tab:avg_speedup}
\vspace{-0.3cm}
\resizebox{.83\columnwidth}{!}{%
\begin{tabular}{@{}lccc@{}}
\toprule
\textbf{Dataset}          & CIFAR-10     & GSC          & Tiny ImageNet \\ \midrule
\textbf{Average speed-up} & 3.9$\times$ & 2.7$\times$ & 3.1$\times$  \\ \bottomrule
\end{tabular}%
}
\vspace{-0.3cm}
\end{table}

\begin{table*}[]
\centering
\caption{Performance on CIFAR-10 of models trained with the MPIC or the NE16 regularizer, and of three baselines quantized at fixed-precision}
\vspace{-0.3cm}
\begin{tabular}{@{}lccccccc@{}}
\hline
\textbf{}    & \multicolumn{1}{l}{\textbf{}}    & \multicolumn{1}{l}{\textbf{}} & \multicolumn{3}{c}{\textbf{MPIC}} & \multicolumn{2}{c}{\textbf{NE16}}    \\ \cmidrule(l){4-6} \cmidrule(l){7-8}
\textbf{$\text{Model}_\text{Training target HW}$} & \multicolumn{1}{l}{\textbf{Test Accuracy (\%)}} & \multicolumn{1}{l}{\textbf{Size (kB)}} & \multicolumn{1}{l}{\textbf{Cycles (\boldmath$\times 10^6$)}} & \multicolumn{1}{l}{\textbf{Latency (ms)}} & \multicolumn{1}{l}{\textbf{Energy (\boldmath$\mu$J)}} & \multicolumn{1}{l}{\textbf{Cycles (\boldmath$\times 10^3$)}} & \multicolumn{1}{l}{\textbf{Latency (ms)}} \\ \hline
$\text{High}_\text{MPIC}$  & 86.28 & 74.98   & 5.038   & 20.15      & 108.46 & 186.014   &  0.50    \\
$\text{Medium}_\text{MPIC}$ & 84.83 & 59.18   & 3.528   & 14.11      & 75.96  & 163.829   &   0.44   \\
$\text{Low}_\text{MPIC}$  & 71.78 & 6.84    & 0.636   & 2.54 & 13.69  & 96.394  &  0.26    \\
$\text{High}_\text{NE16}$  & 86.20 & 75.92   & 5.251   & 21.00    & 113.05 & 149.796   &    0.40  \\
$\text{Medium}_\text{NE16}$ & 84.02 & 47.12   & 2.862   & 11.45 & 61.62  & 117.144   & 0.32     \\
$\text{Low}_\text{NE16}$  & 73.59 & 12.90    & 0.540   & 2.16 & 12.01  & 39.119  &  0.11    \\
w8a8  & 86.26 & 77.36   & 5.953   & 23.81      & 128.17 & 155.241   & 0.42    \\
w4a8  & 85.46 & 38.68   & 5.435   & 21.74      & 117.03 & 104.361   & 0.28    \\
w2a8  & 71.81 & 19.34   & 5.001   & 20.00      & 107.66 & 78.921  & 0.21
     \\

 \hline
\end{tabular}
\label{tab:deployment_results}
  \vspace{-0.3cm}
\end{table*}

In the leftmost plot of Fig.~\ref{fig:icl_mpic_ne16}, we present the accuracy vs. execution cycles trade-off of each model, when deployed on the MPIC hardware.
The rightmost plot presents instead the results obtained by deploying the trained models on the NE16 accelerator. We also report the results of models obtained with the ``wrong'' cost model for both targets, to show the importance of hardware-awareness.
Indeed, while using a cost model tailored to a different hardware does not impact the results significantly in the MPIC case, mainly thanks to the flexibility of this CPU-based platform, the benefit of matching the complexity metric to the actual hardware platform are significant in the case of NE16.
For example, at iso-accuracy with the baseline w2a8, we obtain a model with a 22.14\% increase and a 50.43\% decrease in number of cycles on the NE16 hardware when using MPIC and NE16 cost models, respectively. 
This is due to the more complex dataflow and spacial parallelism of this accelerator, which make it more convenient to assign the same bit-width to \textit{entire chunks of channels}, otherwise some computational bandwidth is wasted and the latency gain is reduced. This is taken into account by the NE16 model, but not by the MPIC one, thus leading to suboptimal solutions.

We report a detailed overview of models sampled from the Pareto curves of Fig.~\ref{fig:icl_mpic_ne16} in Table~\ref{tab:deployment_results}. In particular, we select for each target hardware the Pareto-optimal model with most cycles ($\text{High}$), and the fastest one achieving a validation accuracy greater 70\% ($\text{Low}$). Additionally, we selected a third intermediate model ($\text{Medium}$) as the one with the number of cycles closest to the average between High and the Low. We report the accuracy, size, number of cycles, latency and energy per inference. We also include the baseline fixed-precision models.

The $\text{High}_{\text{MPIC}}$ model has a comparable cost, in terms of latency and energy, with respect to the reference w2a8 on MPIC, but an additional 14.47\% in test accuracy. This is associated to the application of pruning to cut out certain channels of the DNN. However, the same model deployed on the NE16 accelerator is associated to a 136\% increase in latency, because of the mismatch between the two cost models. 
The $\text{Low}_{\text{NE16}}$ model achieves +1.78\% test accuracy and a latency reduction of 89.20\% and 50.43\% on MPIC and NE16, respectively, when compared to the w2a8 baseline.

Notably, we expect the cost model's impact to be particularly evident on \textit{tiny} DNNs. In fact, it is for these models that a precision assignment not fully matched with the target hardware characteristics (e.g. its parallelism) incurs the largest (relative) overhead. Thus, while our method can scale to larger networks and dataset, its HW-awareness feature is particularly relevant at small scale.

\subsection{Ablation studies}
\subsubsection{Models analysis}
\label{sec:results_model_analysis}
\begin{figure*}[t]
  \centering
  \includegraphics[width=.8\textwidth]{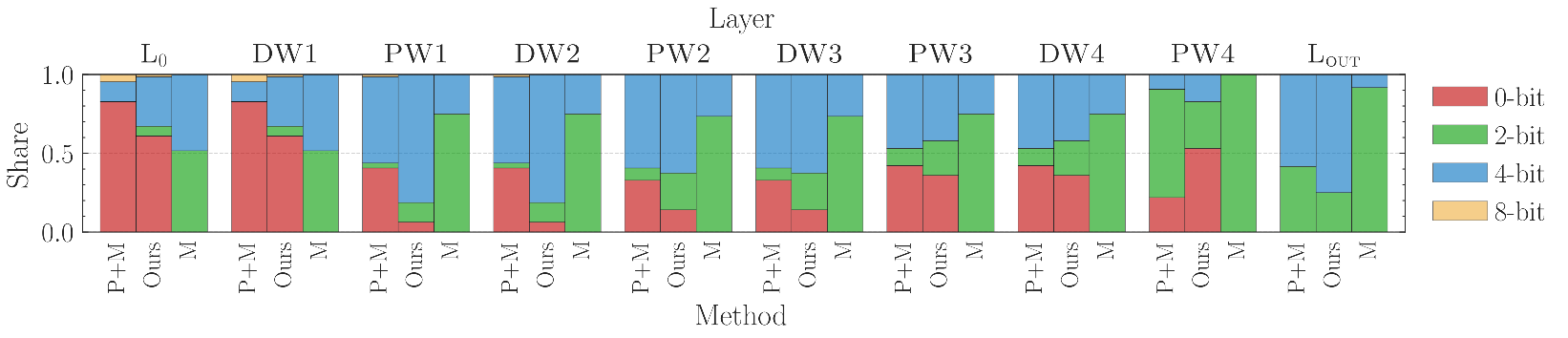}
  \vspace{-0.5cm}
  \caption{Weights channels' bit-width distribution for three models trained with PIT + MixPrec (P+M), joint MixPrec and Pruning (Ours), and MixPrec (M) on GSC with the size regularizer. Input/output layers are denoted as $\textsc{L}_0/\textsc{L}_{\textsc{out}}$, and DW/PW stand for depthwise/pointwise layers, respectively.}
  \vspace{-0.1cm}
\label{fig:model_analysis_perlayer}
\end{figure*}

In Fig.~\ref{fig:model_analysis_perlayer} we report the share of assigned precisions to the weights' channels of each layer. We consider three models trained on the GSC benchmark using the size cost model, comparing our method with MixPrec and MixPrec + PIT. We picked from the Pareto fronts of Fig.~\ref{fig:sota_comparison} the points highlighted with red circles, to analyze the condition in which the different techniques' results differ the most.  
As expected, the concatenation of PIT and MixPrec leads to a higher percentage of pruned channels with respect to our proposed method for almost all the layers. This happens because PIT can only remove entire channels to reduce the DNN's complexity, thus the search space is significantly smaller. The channels which are not pruned are then quantized at a high bit-width by the MixPrec approach not to lose additional capacity. 
On the other hand, our proposed approach tends to apply pruning less and exploit lower bit-widths to reduce the DNN's complexity. This is enabled by the larger search space, which is not restricted by optimization choices of another approach applied beforehand.
The single application of MixPrec, not coupled with any pruning technique, leads to DNNs which have a higher percentage of parameters quantized at 2 bits. This is expected, since for high strength values, 2-bit is the lowest complexity option available for MixPrec, which selects it for most channels, although many of those could be completely eliminated without affecting accuracy (as shown by our method).

\begin{figure*}[t]
  \centering
  \includegraphics[width=1\textwidth]{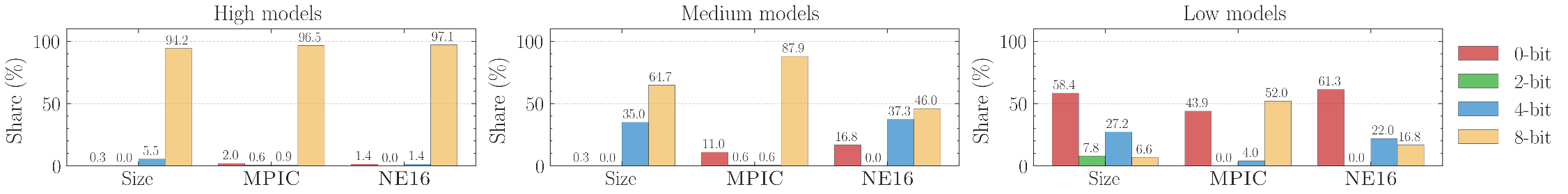}
  \vspace{-0.8cm}
  \caption{Weights bit-widths distributions for models varying in complexity for different regularizers employed in the training objective on CIFAR-10.}
\label{fig:comparisons_large_medium_small}
  \vspace{-0.4cm}
\end{figure*}

In Fig.~\ref{fig:comparisons_large_medium_small} we show the impact of the adopted cost regularizer on the weights precision assignment. We consider three cost models, i.e., Size, MPIC and NE16, with the activations fixed at 8 bits for a fair comparison. We then take three DNNs for each cost model, selected from the Pareto fronts on CIFAR-10 using the same rationale of Table~\ref{tab:deployment_results}.
For all the three "High" DNNs (leftmost plot), most of the weight parameters are quantized at 8 bits, i.e. the highest possibile precision, as expected. 
In "Medium" models (central plot), very few parameters, if any, are quantized at 2 bits, since it is the least interesting precision due to the low representational capacity, but not negligible cost. 
In contrast, 4-bit channels represent a significant component, especially for the size and NE16 regularizers, costing less then twice the 2-bit ones while providing significant accuracy advantages.
For what concerns ``Low'' models, except when trained with the MPIC regularizer, the 4-bit precision is assigned more often than 8-bit. In particular, with a more deepened analysis, it is possible to see that the 8-bit precision is favored only in the final layer, which is a common finding~\cite{cai2020edmips, Yang2020FracBitsMP}. 
Size is the only cost regularizer leading to some weights' channels being quantized at 2 bits. The MPIC cost model mainly favors pruning and keeps most of the other weights at 8 bits, since there is not a sufficient cost difference between this and smaller bit-widths. The NE16 cost model, instead, encourages a more spread-out distribution between 4- and 8-bit but entirely avoids 2-bit precision. 
The reason is that the NE16 cost model does not scale linearly with the output channels, as each processing element (PE) handles groups of 32 output channels (Sec.~\ref{sec:ne16}). Consequently, running a single channel at one precision incurs the same cost as running 32 channels, implying that to enhance latency, the NE16 accelerator should execute at least 32 2-bit filters. However, this would lead to suboptimal solutions from an accuracy standpoint, and is thus avoided by the optimization.

\subsubsection{Impact of activations' quantization}
\label{sec:impact_act_quant}
\begin{figure}[]
  \centering
  \includegraphics[width=\columnwidth]{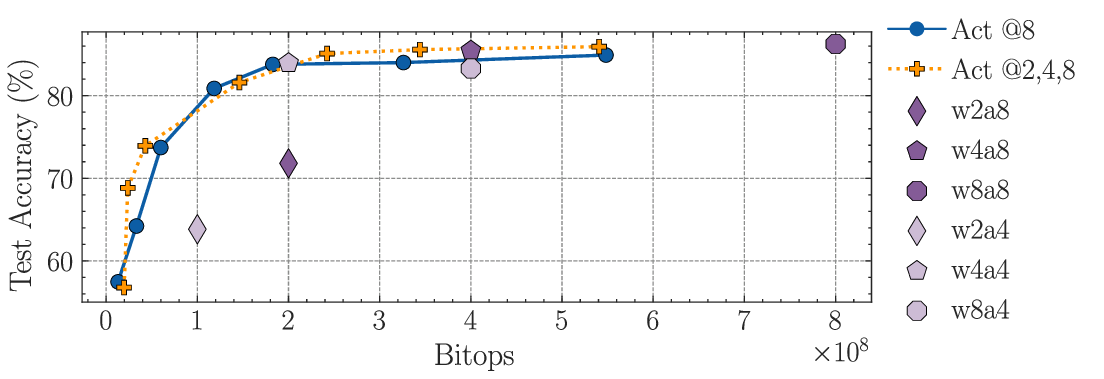}
  \vspace{-0.8cm}
      \caption{Pareto fronts of architectures obtained by either quantizing the activations at 8 bits or assigning the precision layer-wise from the set $P_X = \{2,4,8\}$ on CIFAR-10}
  \vspace{-0.3cm}
  \label{fig:icl_bitops}
\end{figure}
All previous results have been obtained fixing the activations' precision at 8-bit, either because reducing it was not beneficial (when optimizing for model size) or to enable a fair comparison between MPIC and NE16 cost models. Fig.~\ref{fig:icl_bitops} shows instead the result of applying our MPS method also to activations, with a layer-wise granularity, selecting between 2, 4 and 8-bit precision. 
The resulting Pareto curve, in orange, is compared against weights-only MPS with 8-bit activations (in blue). All fixed-precision baselines are also reported, except for the w*a2 ones, which incur a very low accuracy  ($<40\%$) that is not acceptable for any computer vision task.
For the sake of space, we only report the results on CIFAR-10, and we use the bitops cost model~\cite{cai2020edmips} as a hardware-agnostic latency proxy.

Reducing the activations' precision below 8-bit generally provides a better trade-off, as expected. The maximum gain is obtained considering the second point from the left of the two curves: in this case, when allowed to reduce activations' precision, our method improves accuracy by 4.61\% while also reducing the bitops by 28.51\%. However, for all other cases, the average accuracy difference between DNNs with similar bitops is less than 1\%.
This is in contrast with the results of the baselines, where for instance, w4a4 achieves a very good trade-off, with equal bitops and +12.14\% test accuracy compared to w2a8. The differences are also lower than those reported by some previous MPS methods~\cite{cai2020edmips,risso2022igsc}. We argue that this is an effect of the possibility of pruning weight channels introduced by our method. When this option is not available, reducing activations' precision can be highly beneficial; vice versa, when possible, reducing the number of features by pruning weight channels (while keeping their precision at 8-bit) results in a representation capacity that is comparable to keeping all features, but at a reduced bit-width. Insights such as this, which clearly depend on the selected precision set, DNN model, and quantization scheme, can be revealed by our proposed method, and might be useful to drive future hardware design decisions.

\section{Conclusion}
We have proposed a gradient-based optimization method for DNNs that is able to apply structured pruning and channel-wise MPS at the same time. This enables a speed-up in the optimization process, since it avoids the time-consuming sequential application of pruning and quantization techniques. Furthermore, it allows the exploration of a broader search space, not restricted by the choices of the first optimization technique applied. In addition, we have shown the importance of hardware-aware cost models for this optimization.
Our method is able to achieve up to 56.17\% size reduction at iso-accuracy when compared to a previous state-of-the-art method. With respect to 8-bit and 2-bit fixed-precision DNNs, we obtain a size reduction of up to 47.50\% and 69.54\%, respectively, without accuracy drops.
As future work directions, we would like to expand our approach to support also transformer-based architectures.
 \label{sec:conclusion}

\ifCLASSOPTIONcaptionsoff
  \newpage
\fi

\bibliographystyle{IEEEtran}
\bibliography{bstctl,references}

\begin{thebibliography}{10}
\providecommand{\url}[1]{#1}
\csname url@samestyle\endcsname
\providecommand{\newblock}{\relax}
\providecommand{\bibinfo}[2]{#2}
\providecommand{\BIBentrySTDinterwordspacing}{\spaceskip=0pt\relax}
\providecommand{\BIBentryALTinterwordstretchfactor}{4}
\providecommand{\BIBentryALTinterwordspacing}{\spaceskip=\fontdimen2\font plus
\BIBentryALTinterwordstretchfactor\fontdimen3\font minus \fontdimen4\font\relax}
\providecommand{\BIBforeignlanguage}[2]{{%
\expandafter\ifx\csname l@#1\endcsname\relax
\typeout{** WARNING: IEEEtran.bst: No hyphenation pattern has been}%
\typeout{** loaded for the language `#1'. Using the pattern for}%
\typeout{** the default language instead.}%
\else
\language=\csname l@#1\endcsname
\fi
#2}}
\providecommand{\BIBdecl}{\relax}
\BIBdecl

\bibitem{liberis2021micronas}
E.~Liberis, L.~Dudziak, and N.~D. Lane, ``$\mu$nas: Constrained neural architecture search for microcontrollers,'' in \emph{Proc. 1st Workshop Mach. Learn. Syst.}, 2021, p. 70–79.

\bibitem{han2015efficient}
S.~Han \emph{et~al.}, ``Learning both weights and connections for efficient neural networks,'' in \emph{Proc. 28th Int. Conf. Neural Inf. Proc. Syst.}, vol.~1, 2015, pp. 1135--1143.

\bibitem{jacob2018quantization}
B.~Jacob \emph{et~al.}, ``Quantization and training of neural networks for efficient integer-arithmetic-only inference,'' in \emph{Proc. IEEE/CVF Conf. Comput. Vis. Pattern Recogn.}, 2018, pp. 2704--2713.

\bibitem{white2023nas1000papers}
C.~White \emph{et~al.}, ``Neural architecture search: Insights from 1000 papers,'' \emph{arXiv:2301.08727}, 2023.

\bibitem{benmeziane2021hwawarenas}
H.~Benmeziane \emph{et~al.}, ``Hardware-aware neural architecture search: Survey and taxonomy,'' in \emph{Proc. 30th Int. Joint Conf. Artif. Intell.}, 2021, pp. 4322--4329.

\bibitem{risso2023pit}
M.~Risso \emph{et~al.}, ``Lightweight neural architecture search for temporal convolutional networks at the edge,'' \emph{IEEE Trans. Comput.}, vol.~72, no.~3, pp. 744--758, 2023.

\bibitem{cai2020edmips}
Z.~Cai and N.~Vasconcelos, ``Rethinking differentiable search for mixed-precision neural networks,'' in \emph{Proc. IEEE/CVF Conf. Comput. Vis. Pattern Recogn.}, 2020, pp. 2346--2355.

\bibitem{risso2022igsc}
M.~Risso \emph{et~al.}, ``Channel-wise mixed-precision assignment for dnn inference on constrained edge nodes,'' in \emph{Proc. IEEE 13th Int. Green Sust. Comp. Conf.}, 2022, pp. 1--6.

\bibitem{ottavi2020mpic}
G.~Ottavi \emph{et~al.}, ``A mixed-precision risc-v processor for extreme-edge dnn inference,'' in \emph{Proc. IEEE Comput. Soc. Ann. Symp. on VLSI}, 2020, pp. 512--517.

\bibitem{macan2023ne16}
L.~Macan \emph{et~al.}, ``Wip: Automatic dnn deployment on heterogeneous platforms: the gap9 case study,'' in \emph{Proc. Int. Conf. on Compil., Arch., Synth. for Emb. Syst.}, 2023, pp. 9--10.

\bibitem{google-speech-commands}
P.~Warden, ``Speech commands: A dataset for limited-vocabulary speech recognition,'' \emph{arXiv:1804.03209}, 2018.

\bibitem{dai2021vsquant}
S.~Dai \emph{et~al.}, ``Vs-quant: Per-vector scaled quantization for accurate low-precision neural network inference,'' in \emph{Proc. Mach. Learn. Syst.}, vol.~3, 2021, pp. 873--884.

\bibitem{frantar2022gptq}
E.~Frantar \emph{et~al.}, ``{GPTQ}: Accurate post-training compression for generative pretrained transformers,'' \emph{arXiv:2210.17323}, 2022.

\bibitem{choi2018pact}
J.~Choi \emph{et~al.}, ``Pact: Parameterized clipping activation for quantized neural networks,'' \emph{arXiv:1805.06085}, 2018.

\bibitem{lq_nets}
D.~Zhang \emph{et~al.}, ``Lq-nets: Learned quantization for highly accurate and compact deep neural networks,'' in \emph{Proc. Europ. Conf. Comput. Vis.}, 2018, pp. 365--382.

\bibitem{nas_rl}
B.~Zoph and Q.~V. Le, ``Neural architecture search with reinforcement learning,'' \emph{arXiv:1611.01578}, 2016.

\bibitem{liu2018darts}
H.~Liu, K.~Simonyan, and Y.~Yang, ``Darts: Differentiable architecture search,'' \emph{arXiv:1806.09055}, 2018.

\bibitem{haq}
K.~Wang \emph{et~al.}, ``Haq: Hardware-aware automated quantization with mixed precision,'' in \emph{Proc. IEEE/CVF Conf. Comput. Vis. Pattern Recogn.}, 2019.

\bibitem{releq}
A.~T. Elthakeb \emph{et~al.}, ``Releq : A reinforcement learning approach for automatic deep quantization of neural networks,'' \emph{IEEE Micro}, vol.~40, no.~5, pp. 37--45, 2020.

\bibitem{tan2019mnasnet}
M.~Tan \emph{et~al.}, ``Mnasnet: Platform-aware neural architecture search for mobile,'' in \emph{Proc. IEEE/CVF Conf. Comput. Vis. Pattern Recogn.}, 2019, pp. 2820--2828.

\bibitem{pruning_survey}
T.~Hoefler \emph{et~al.}, ``Sparsity in deep learning: Pruning and growth for efficient inference and training in neural networks,'' \emph{Journ. Mach. Learn. Res.}, vol.~22, no.~1, 2021.

\bibitem{deep_compression}
S.~Han, H.~Mao, and W.~J. Dally, ``Deep compression: Compressing deep neural networks with pruning, trained quantization and huffman coding,'' in \emph{Int. Conf. Learn. Repr.}, 2016.

\bibitem{scalpel}
J.~Yu \emph{et~al.}, ``Scalpel: Customizing dnn pruning to the underlying hardware parallelism,'' in \emph{Proc. ACM/IEEE 44th Ann. Int. Symp. Comput. Arch.}, 2017, pp. 548--560.

\bibitem{data_free_struct}
H.~Li \emph{et~al.}, ``Pruning filters for efficient convnets,'' in \emph{Int. Conf. Learn. Repr.}, 2017.

\bibitem{thinet}
J.~Luo, J.~Wu, and W.~Lin, ``Thinet: A filter level pruning method for deep neural network compression,'' in \emph{Proc. IEEE Int. Conf. Comput. Vis.}, 2017, pp. 5068--5076.

\bibitem{morphnet}
A.~Gordon \emph{et~al.}, ``Morphnet: Fast \& simple resource-constrained structure learning of deep networks,'' in \emph{Proc. IEEE/CVF Conf. Comput. Vis. Pattern Recogn.}, 2018, pp. 1586--1595.

\bibitem{mixed_darts}
B.~Wu \emph{et~al.}, ``Mixed precision quantization of convnets via differentiable neural architecture search,'' \emph{arXiv:1812.00090}, 2018.

\bibitem{Yang2020FracBitsMP}
L.~Yang and Q.~Jin, ``Fracbits: Mixed precision quantization via fractional bit-widths,'' in \emph{Proc. AAAI Conf. Artif. Intell.}, 2021.

\bibitem{Lou2020AutoQ:}
Q.~Lou \emph{et~al.}, ``Autoq: Automated kernel-wise neural network quantization,'' in \emph{Int. Conf. Learn. Repr.}, 2020.

\bibitem{fbnet_energy}
C.~Gong \emph{et~al.}, ``Mixed precision neural architecture search for energy efficient deep learning,'' in \emph{Proc. IEEE/ACM Int. Conf. Comput.-Aid. Design}, 2019, pp. 1--7.

\bibitem{bayesian_bits}
M.~Van~Baalen \emph{et~al.}, ``Bayesian bits: Unifying quantization and pruning,'' in \emph{Proc. 33th Int. Conf. Neural Inf. Proc. Syst.}, vol.~33, 2020, pp. 5741--5752.

\bibitem{diff_joint_pq}
Y.~Wang, Y.~Lu, and T.~Blankevoort, ``Differentiable joint pruning and quantization for hardware efficiency,'' in \emph{Proc. Europ. Conf. Comput. Vis.}, 2020, pp. 259--277.

\bibitem{Chitty-Venkata2022}
K.~T. Chitty-Venkata \emph{et~al.}, ``Efficient design space exploration for sparse mixed precision neural architectures,'' in \emph{Proc. 31st Int. Symp. High-Perf. Parall. Distrib. Comp.}, 2022, p. 265–276.

\bibitem{hawq_v2}
Z.~Dong \emph{et~al.}, ``Hawq-v2: Hessian aware trace-weighted quantization of neural networks,'' in \emph{Proc. 33th Int. Conf. Neural Inf. Proc. Syst.}, 2020, pp. 18\,518--18\,529.

\bibitem{sqnr_mps}
D.~Lin, S.~Talathi, and S.~Annapureddy, ``Fixed point quantization of deep convolutional networks,'' in \emph{Proc. 33rd Int. Conf. Mach. Learn.}, 2016, pp. 2849--2858.

\bibitem{bp-nas}
H.~Yu \emph{et~al.}, ``Search what you want: Barrier panelty nas for mixed precision quantization,'' in \emph{Proc. Europ. Conf. Comput. Vis.}, 2020, pp. 1--16.

\bibitem{free_bits}
G.~Rutishauser, F.~Conti, and L.~Benini, ``Free bits: Latency optimization of mixed-precision quantized neural networks on the edge,'' in \emph{Proc. IEEE 5th Int. Conf. Artif. Intell. Circ. Syst.}, 2023, pp. 1--5.

\bibitem{Wang2020APQ}
T.~Wang \emph{et~al.}, ``Apq: Joint search for network architecture, pruning and quantization policy,'' in \emph{Proc. IEEE/CVF Conf. Comput. Vis. Patt. Recogn.}, 2020, pp. 2075--2084.

\bibitem{Chitty-Venkata2023}
K.~T. Chitty-Venkata \emph{et~al.}, ``Differentiable neural architecture, mixed precision and accelerator co-search,'' \emph{IEEE Access}, vol.~11, pp. 106\,670--106\,687, 2023.

\bibitem{chitty-venkata_neural_2023}
K.~T. Chitty-Venkata and A.~K. Somani, ``\BIBforeignlanguage{en}{Neural {Architecture} {Search} {Survey}: {A} {Hardware} {Perspective}},'' \emph{\BIBforeignlanguage{en}{ACM Comput. Surv.}}, vol.~55, no.~4, pp. 1--36, Apr. 2023.

\bibitem{cai_enable_2022}
H.~Cai \emph{et~al.}, ``\BIBforeignlanguage{en}{Enable deep learning on mobile devices: Methods, systems, and applications},'' \emph{\BIBforeignlanguage{en}{ACM Trans. Design Autom. Electr. Syst.}}, vol.~27, no.~3, pp. 1--50, May 2022.

\bibitem{prasad_specialization_2023}
A.~S. Prasad, L.~Benini, and F.~Conti, ``\BIBforeignlanguage{en}{Specialization meets flexibility: a heterogeneous architecture for high-efficiency, high-flexibility {AR}/{VR} processing},'' in \emph{\BIBforeignlanguage{en}{Proc. ACM/IEEE Design Autom. Conf.}}, 2023, pp. 1--6.

\bibitem{plinio_fdl}
D.~J. Pagliari \emph{et~al.}, ``Plinio: A user-friendly library of gradient-based methods for complexity-aware dnn optimization,'' in \emph{Proc. Forum Specif. Design Lang.}, 2023, pp. 1--8.

\bibitem{mlperftiny-benchmark}
C.~Banbury \emph{et~al.}, ``Mlperf tiny benchmark,'' in \emph{Proc. Neural Inf. Proc. Syst. Track on Datasets and Benchmarks}, 2021.

\bibitem{wan2020fbnetv2}
A.~Wan \emph{et~al.}, ``Fbnetv2: Differentiable neural architecture search for spatial and channel dimensions,'' in \emph{Proc. IEEE/CVF Conf. Comput. Vis. Pattern Recogn.}, 2020, pp. 12\,965--12\,974.

\end{thebibliography}

\end{document}